\documentclass[10pt,journal,compsoc]{IEEEtran}

\usepackage{amsmath,amsfonts}
\usepackage{algorithmic}
\usepackage{graphicx}
\usepackage{textcomp}
\usepackage{xcolor}
\usepackage{xcolor}
\usepackage{xspace}
\usepackage[ruled, vlined, linesnumbered]{algorithm2e}
\usepackage{algorithmic}
\usepackage{graphicx}
\usepackage{amsmath}
\usepackage{amsfonts}
\usepackage{subfig}
\usepackage{amsthm}
\usepackage{mathrsfs}
\usepackage{booktabs}
\usepackage{color}
\usepackage{hyperref}
\usepackage{lipsum}
\usepackage{subfig}
\usepackage{paralist}
\usepackage{balance}
\usepackage[numbers]{natbib}

\newcommand{\ourm}{\textsc{ReVAMP}\xspace}
\newcommand{\tel}{Shanghai-Telecom\xspace}
\newcommand{\tdk}{TalkingData\xspace}
\newcommand{\cin}{check-in\xspace}
\newcommand{\cins}{check-ins\xspace}

\newcommand{\eg}{\emph{e.g.}}
\newcommand{\ie}{\emph{i.e.}}
\newcommand{\etc}{\emph{etc.}}

\theoremstyle{definition}
\newtheorem{definition}{Definition}

\newtheorem*{problem*}{Problem Statement}

\newcommand{\cat}[1]{`\textit{#1}'}
\newcommand{\sap}[1]{smartphone-app}
\newcommand{\cm}[1]{\mathcal{#1}}

\newcommand{\bs}[1]{\boldsymbol{#1}}

\usepackage{colortbl}
\usepackage{titlesec}

\newcommand{\xhdr}[1]{\vspace{0mm}\noindent{{\bf #1.}}}

\titlespacing\section{0pt}{12pt plus 4pt minus 2pt}{0pt plus 2pt minus 2pt}
\titlespacing\subsection{0pt}{8pt plus 2pt minus 1pt}{0pt plus 2pt minus 2pt}
\begin{document}
\title{Modeling Spatial Trajectories using Coarse-Grained Smartphone Logs}

\author{Vinayak~Gupta and~Srikanta~Bedathur
\IEEEcompsocitemizethanks{\IEEEcompsocthanksitem Vinayak Gupta and Srikanta Bedathur are with the Department of Computer
Science \& Engineering, Bharti Building, Indian Institute of Technology, Delhi, Hauz Khas, New Delhi, 110015, India. \protect\\
E-mail: \{vinayak.gupta, srikanta\}@cse.iitd.ac.in}
\thanks{Manuscript received XX XX, XXXX.}
}

\markboth{IEEE TRANSACTIONS ON BIG DATA,~Vol.~XX, No.~X, August~2022}%
{Shell \MakeLowercase{\textit{et al.}}: Bare Demo of IEEEtran.cls for Computer Society Journals}
\IEEEtitleabstractindextext{
\begin{abstract}
Current approaches for points-of-interest (POI) recommendation learn the preferences of a user via the standard spatial features such as the POI coordinates, the social network, \etc\ These models ignore a crucial aspect of spatial mobility -- \textit{every user carries their smartphones wherever they go}. In addition, with growing privacy concerns, users refrain from sharing their exact geographical coordinates and their social media activity. In this paper, we present \ourm, a sequential POI recommendation approach that utilizes the user activity on smartphone applications (or apps) to identify their mobility preferences. This work aligns with the recent psychological studies of online urban users, which show that their spatial mobility behavior is largely influenced by the activity of their smartphone apps. In addition, our proposal of \textit{coarse-grained} smartphone data refers to data logs collected in a privacy-conscious manner, \ie, consisting only of (a) category of the smartphone app and (b) category of \cin location. Thus, \ourm is not privy to precise geo-coordinates, social networks, or the specific application being accessed. Buoyed by the efficacy of self-attention models, we learn the POI preferences of a user using two forms of positional encodings -- absolute and relative -- with each extracted from the inter-\cin dynamics in the \cin sequence of a user. Extensive experiments across two large-scale datasets from China show the predictive prowess of \ourm and its ability to predict app- and POI categories.
\end{abstract}

\begin{IEEEkeywords}
Sequential Recommendation; Smartphone Apps; Spatial and Temporal Data; Self-Attention;
\end{IEEEkeywords}}

\maketitle
\IEEEdisplaynontitleabstractindextext
\IEEEpeerreviewmaketitle

\IEEEraisesectionheading{\section{Introduction}\label{sec:introduction}}
\IEEEPARstart{R}{apid} advancements in the smartphone industry and ubiquitous internet access have led to an exponential growth in the number of available users and internet-based applications. Moreover, smartphones have become increasingly prevalent with up to 345 million units sold in the first quarter of 2021\footnote{\url{https://bit.ly/3wyPu8Y} (Accessed July 2022)}. As a result, the online footprint of a user spans multiple applications, with an average smartphone owner accessing 10 smartphone applications (or apps) every day and 30 apps each month\footnote{\url{https://buildfire.com/app-statistics/} (Accessed July 2022)}. These footprints can be perceived as the digitized nature of the user's proclivity in different domains and can be collected without any personally identifiable information (PII)\footnote{\url{https://w.wiki/5d5C} (Accessed July 2022)}. Thus, maintaining the privacy of a user~\cite{cross, silkroad, li}. Recent research~\cite{revisit, revisit2} has shown that the online web activity of a user exhibits \textit{revisitation} patterns, \ie, a user is likely to visit certain apps repetitively with similar time intervals between corresponding visits. \citet{remob} and~\citet{wwwmob} have shown that these online patterns are analogous to the user's spatial mobility preferences, \ie, the current geographical location can influence the web-browsing activities of a user.  

\begin{figure}[b]
\centering
  \includegraphics[width=\columnwidth]{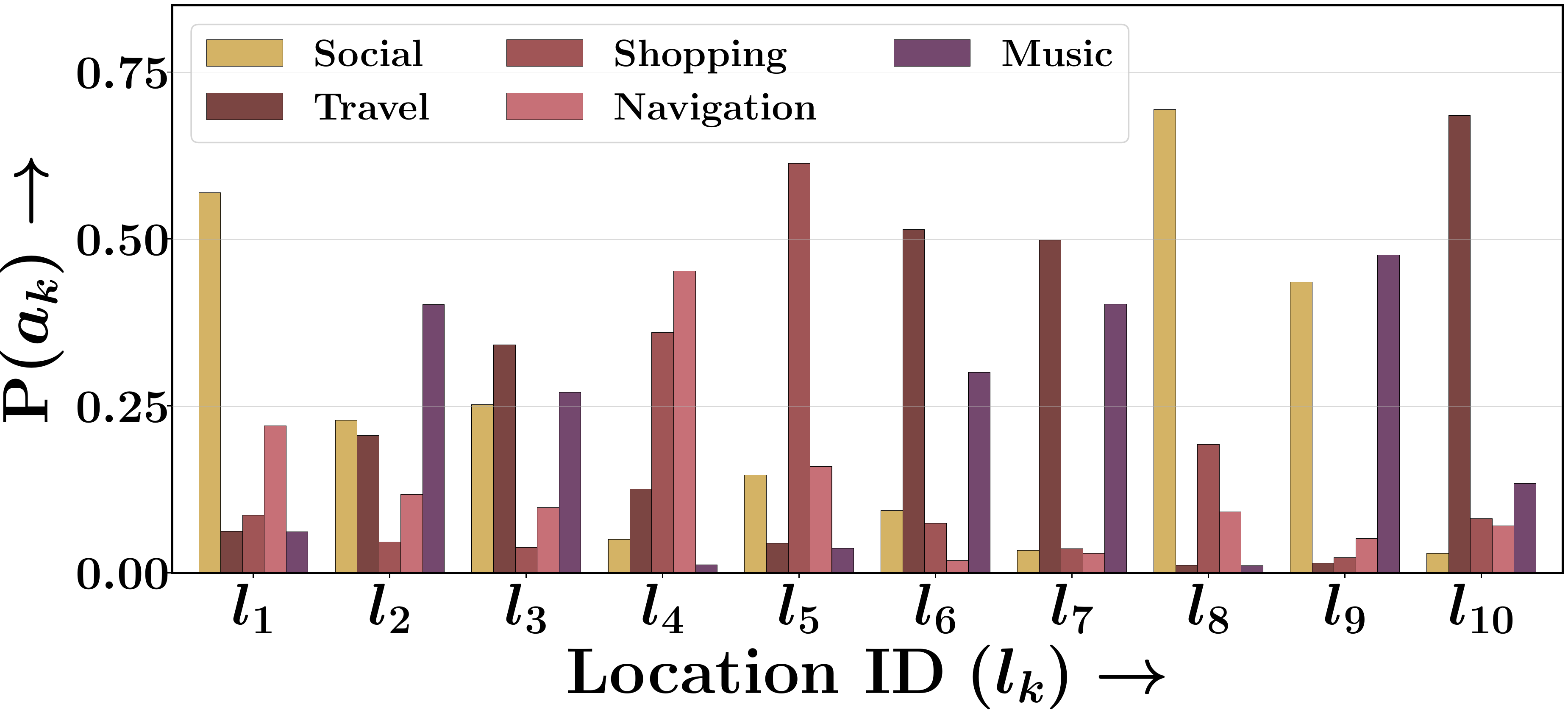}
  \vspace{-5mm}
\caption{The probability of a smartphone-app category --  among \cat{Social}, \cat{Travel}, \cat{Shopping}, \cat{Navigation}, and \cat{Music} -- to be used at 10 most popular locations from \tel dataset. The plot indicates that the smartphone app-usage depends on the \cin location.}
\label{fig:mdl_tel_cat}
\end{figure}

\xhdr{Limitations of Prior Works}
Modern POI recommendation approaches~\cite{deepmove, axolotl, lbsn2vec} utilize the standard features specific to a user and a POI -- social network, geo-coordinates, and category classifications of POI -- to learn the mobility patterns of a user. The situation has been exacerbated in recent times due to the advent of restrictions on personal data collection and a growing awareness (in some geopolitical regions) about the need for personal privacy~\cite{privacy, privacy2}. Moreover, current approaches overlook two crucial aspects of urban computing -- the exponential growth of online platforms and the widespread use of smartphones. Undeniably, everyone carries and simultaneously uses a smartphone wherever they go. Thus, standard approaches are inappropriate to design POI recommender systems that must capture the location influence on the apps used by a user. To highlight the importance of the POI-app relationship, we plot the category of the app used by all users at the ten most popular locations from our \tel dataset~\cite{li} in Figure~\ref{fig:mdl_tel_cat}. The plot shows that the \cin locations can influence a user to visit apps of certain categories more than other apps. We note that this POI influence over the category of the app being accessed is similar across different users. 

The correlation between spatial mobility and smartphone use is essential to address the problems related to user demographics~\cite{carat, f2f2}, trajectory analysis~\cite{coarse}, app recommendation~\cite{appusage2vec}, and to identify hotspots for network operators~\cite{wwwmob}. However, utilizing smartphone usage for sequential POI recommendations is not addressed in the past literature. The papers most similar to our work are by~\cite{reapp2} and~\cite{f2f}. \citet{reapp2} utilizes a Dirichlet process to determine the next user location, but it completely disregards the user's privacy, \ie, requires precise geo-coordinates, and~\citet{f2f} is limited to the cold-start recommendation.

\begin{figure}[t]
\centering
  \includegraphics[width=\columnwidth]{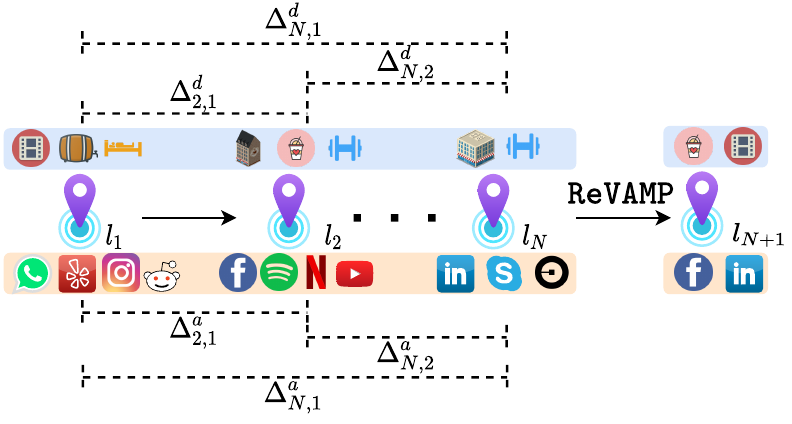}
\vspace{-6mm}
\caption{\ourm learn the dynamics of a \cin sequence via inter-\cin time, app, and POI variations. Here, $\Delta^a_{\bullet, \bullet}$ and $\Delta^d_{\bullet, \bullet}$ denote the smartphone-app and POI-based differences respectively.}
\vspace{-4mm}
\label{fig:mdl_intro}
\end{figure}

\subsection{Present Work}
\noindent In this paper, we present \textbf{\ourm}(\textbf{Re}lative position \textbf{V}ector for \textbf{A}pp-based \textbf{M}obility \textbf{P}rediction), a sequential POI recommendation model that learns the POI and app affinities of smartphone users while simultaneously preserving their privacy. In detail, we consider each \cin as an event involving a smartphone activity, and the physical presence at a POI and learn the correlation between the smartphone-app preferences and the spatial mobility preferences of a user. In addition, to preserve the privacy of a user, we limit our modeling to two aspects of urban mobility: (a) the types of smartphone apps used during a \cin and (b) the category of the \cin location. Thus, the proposed approach is not privy to any PII such as the precise smartphone app being accessed, \eg\ \cat{Facebook}, \cat{Amazon}, \etc; the accurate geo-location, inter-\cin distance, or the social network of a user. Buoyed by the success of self-attention~\cite{transformer} models in sequence modeling, \ourm encodes the dynamic \cin preferences in the user trajectory as a weighted aggregation of all past \cins. 

To better capture the evolving POI and app preferences, it models the variation between each \cin in the sequence using \textit{absolute} and \textit{relative} positional encodings~\cite{postran, reltran}. Specifically, we embed three properties associated with each \cin -- the smartphone app-category, POI-category, and the time of \cin -- and model the temporal evolution as the inter-\cin embedding differences, independently across these features. Figure~\ref{fig:mdl_intro} demonstrates how \ourm embeds and learns the inter-\cin dynamics between the app and POI categories to determine the next POI for a user. Moreover, \ourm can predict the category of the most likely smartphone app to be accessed and the POI category at the next \cin. Such an ability has limitless applications ranging from smartphone app recommendation and bandwidth modeling by cellular network providers~\cite{f2f,f2f2,appusage2vec}. To summarize, the key contributions we make in this paper are three-fold:
\begin{enumerate}[(i)]
\item We propose a self-attention-based approach, called \ourm, to learn the POI preferences of a user via the coarse-grained smartphone usage logs. \ourm returns a ranked list of candidate POIs and the most likely app and POI category for the next \cin.

\item We preserve the privacy needs of a user by learning a personalized sequence encoding for every user. In detail, we force our model to learn the evolving spatial preferences using the variations between each \cin in the sequence based on app category, POI category, and time of the \cin.  Thus, our approach is not privy to accurate geo-locations and social networks. 

\item Exhaustive experiments over two large-scale datasets from China show that \ourm outperforms other state-of-the-art methods for sequential POI recommendation, next app, and location-category prediction tasks. Moreover, we perform a detailed analysis of each component of \ourm, a convergence analysis, and a hyper-parameter analysis to ascertain its practicability. 
\end{enumerate}


\section{Related Work} \label{sec:relwork}
\noindent In this section, we highlight some relevant works to our paper. It mainly falls into three categories -- modeling smartphone and mobility, sequential recommendation, and relative positional encodings for self-attention models.

\xhdr{Modeling Smartphone and Mobility}
Understanding the mobility dynamics of a user has wide applications ranging from location-sensitive advertisements, social community of user, and disease propagation~\cite{cho,colab,coarse}. Traditional mobility prediction models utilized a function-based learning for spatial preferences but were highly susceptible to irregular events in the user trajectory~\cite{cheng2013you}. Therefore, modern approaches~\cite{deepmove,axolotl} utilize a neural network to model the complex user-POI relationships, geographical features, travel distances, and category distribution. These approaches consider the user trajectory as a \cin sequence and train their model parameters by capturing the influences across different sequences. Other approaches~\cite{reformd, attnloc2} include the continuous-time contexts for modeling the time-evolving preferences of a user. However, prior research has shown that users exhibit \textit{revisitation} patterns on their web activities~\cite{revisit, revisit2} and these revisitation patterns resonate with the mobility preferences of a user~\cite{reapp, reapp2}. As per the permissions given by a user to an app, leading corporations, such as Foursquare, utilize smartphone activities to better understand the likes and dislikes of a user to give better POI recommendations~\cite{use2}. The correlation between spatial mobility and smartphone use is essential to address the problems related to user demographics~\cite{carat, f2f2}, trajectory analysis~\cite{coarse}, app recommendation~\cite{appusage2vec}, and to identify hotspots for network operators~\cite{wwwmob}. However, utilizing smartphone usage for sequential POI recommendations is not addressed in the past literature. The papers most similar to our work are by \citet{reapp2} and \citet{f2f}. \citet{reapp2} utilizes a Dirichlet process to determine the next user location, but it completely disregards the user's privacy, \ie, requires precise geo-coordinates, and~\citet{f2f} is limited to \textit{cold-start} POI recommendation rather than sequential recommendations. 

\xhdr{Sequential Recommendation}
Standard collaborative filtering (CF) and matrix factorization (MF) based recommendation approaches~\cite{ncf, reformd} return a list of most likely items that a user will purchase in the future. However, these approaches ignore the temporal context associated with the preferences, \ie, it evolves with time. The task of a sequential recommender system is to continuously model the user-item interactions in the past purchases (or \cins) and predict future interactions. Traditional sequence modeling approaches such as personalized Markov chains~\cite{fpmc} combine matrix factorization with inter-item influences to determine the time-evolving user preferences. However, it has limited expressivity and cannot model complex functions. Neural models such as~\citet{gru4rec, gru4recplus} utilize a recurrent neural network (RNN) to embed the time-conditioned user preferences. Recent research has shown that including attention~\cite{attention} within the RNN architecture achieved better prediction performances even for POI recommendations~\cite{attnloc2, attnloc4}. However, all these approaches were outperformed by the self-attention-based sequential recommendation models~\cite{sasrec}. In detail, \citet{sasrec} embeds user preferences using a weighted aggregation of all past user-item interactions. However, due to largely the heterogeneous nature of data in spatial datasets \eg\ POI category, geographical distance, \etc\ extending such models for sequential POI recommendation is a non-trivial task.

\xhdr{Relative Positional Encodings}
The self-attention models are oblivious of the position of events in the sequence and thus, the original proposal to capture the order of events used fixed function-based encodings~\cite{transformer}. However, recent research on positional encodings~\cite{postran, reltran} has shown that modeling the position as a relative pairwise function between all events of a sequence, in addition to the fixed-function encodings, achieves significant improvements over the standard method. Thus, such relative encodings have been used in a wide range of applications -- primarily for determining the relative word order in natural language tasks~\cite{rel_nlp3,tacl}. Since time is an essential component for learning the dynamics of a time-series~\cite{neuroseqret, imtpp, proactive}, these relative encodings have also been incorporated in item-based recommender systems~\cite{tisasrec} through \textit{time-interval} based inter-event relevance and in POI recommendation~\cite{geosasrec} through geographical distance-based variances. However, the former approach cannot be extended to model the heterogeneous nature of smartphone mobility data and the latter requires precise geographical coordinates.
\section{Problem Formulation} \label{sec:psetup}
\noindent We consider a setting with a set of users as $\cm{U}$ and a set of locations (or POIs), $\cm{L}$. We embed each POI using a $D$ dimensional vector and denote the embedding matrix as $\bs{L} \in \mathbb{R}^{|\cm{L}| \times D}$. We represent the mobile trajectory of a user $u_i$ as a sequence of \cins, $e^{u_i}_k \in \cm{E}^{u_i}$, with each \cin comprising the smartphone app and location details. For a better understanding of our model, let us consider a toy sequence with five \cins to POIs with categories, -- \cat{Bar}, \cat{Cafe}, \cat{Burger-Joint}, \cat{Cafe}, and \cat{Sushi Restaurant}, while using smartphone apps categories -- \cat{Social}, \cat{Shopping}, \cat{Game}, \cat{Social}, and \cat{Travel}, respectively. Thus, for this example, \ourm will use the details of the first four \cins to predict the last \cin.

\begin{definition}[\textsf{Check-ins}]
\label{checkin}
\textit{We define a \cin as a timestamped activity of a user with her smartphone and location details. Specifically, we represent the $k$-th \cin in $\cm{E}$ as $e_k = \left\{l_k, t_k, \cm{A}_k, \cm{S}_k \right\}$ where $l_k$ and $t_k$ denote the POI and \cin time respectively. Here, $\cm{A}_k$ denotes the categories set of the smartphone-app accessed by a user, and $\cm{S}_k$ denotes the set of POI categories.}
\end{definition}
\noindent With a slight abuse of notation, we denote a \cin sequence as $\cm{E}$ and the set of all app- and location categories till a $k$-th \cin as $\cm{A}^*_k = \bigcup_{i=1}^{k} \cm{A}_k$ and $\cm{S}^*_k = \bigcup_{i=1}^{k} \cm{S}_k$ respectively. Now, we formally define the problem of sequential POI recommendations. For our example, $\cm{A}$ will consist of \cat{Shopping}, \cat{Game}, \cat{Social}, and \cat{Travel}, while $\cm{S}$ will include \cat{Bar}, \cat{Burger-Joint}, \cat{Cafe}, and \cat{Sushi Restaurant} respectively.

\begin{problem*}[\textbf{Personalized Sequential Recommendation}]
\label{problem}
\textit{Using the user's past \cin records consisting of app and POI categories, we aim to get a ranked list of the most likely locations the user is expected to visit in her next \cin. Specifically, we learn the time-evolving variation in smartphone and physical mobility to estimate her future preference towards different locations in her vicinity}. 
\end{problem*}
Mathematically, given the first $k$ \cins in a sequence as $\cm{E}_k$, we aim to identify the set of candidate POI for the next \cin, \ie, $e_{k+1}$, conditioned on the app- and location-categories of all \cins in the history. Specifically, maximize the following probability:
\begin{equation}
\mathbb{P}^* = \arg\max_{\Theta} \{ \mathbb{E}[e_{k+1}| \cm{E}_{k}, \cm{A}^*_k, \cm{S}^*_k]\}
\end{equation}
where $\mathbb{E}[e_{k+1}]$ calculates the expectation of $e_{k+1}$ being in the sequence of the user, $\cm{E}_{k}$ given the past \cins of a user. Here, $\Theta$ denotes the \ourm model parameters.

\begin{figure}[t]
\centering
  \includegraphics[height=4cm]{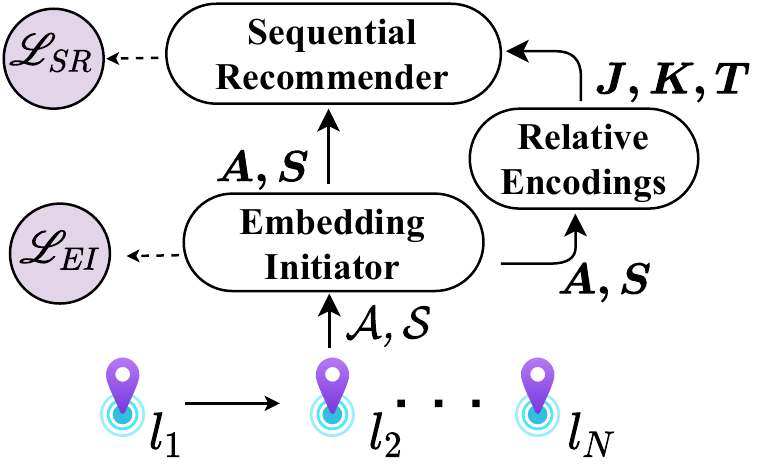}
  \vspace{-2mm}
\caption{Overview of the neural architecture of \ourm.}
\label{fig:overall}
\end{figure}

\begin{figure*}[t!]
\centering
\subfloat[Embedding Initiator (EI)]
{\includegraphics[height=4cm]{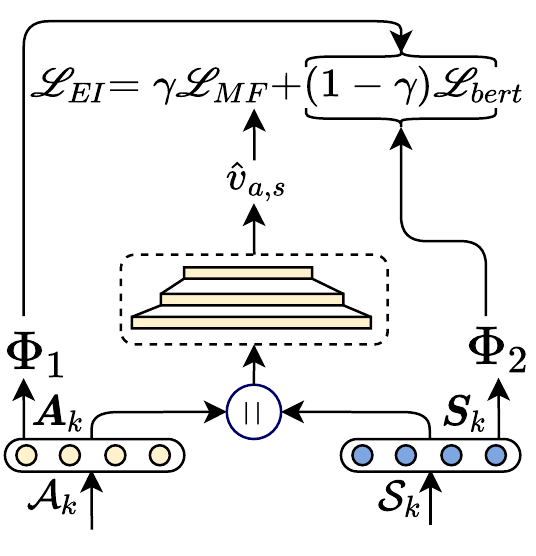}}
\hspace{2cm}
\subfloat[Sequential Recommender (SR)]
{\includegraphics[height=4cm]{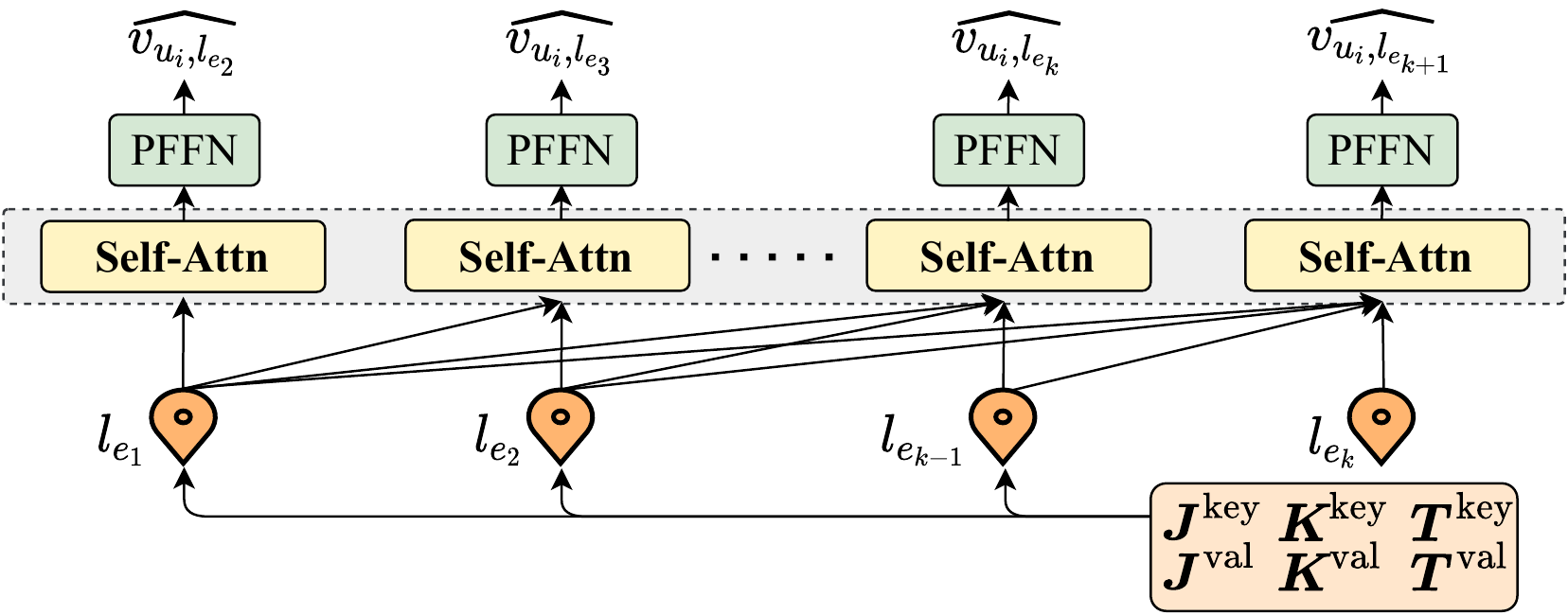}}
\vspace{-2mm}
\caption{Architecture of different components in \ourm.
Panel (a) illustrates the setup of the Embedding Initiator (EI) that learns the category representations.
Panel (b) shows the self-attention architecture in Sequential Recommender (SR).
Note that the input to the self-attention is an aggregation of all past events and relative positional encodings.}
\label{fig:model}
\vspace{-2mm}
\end{figure*}

\section{\ourm Framework}\label{sec:model}
\noindent In this section, we first present a high-level overview of the deep neural network architecture of \ourm and then describe component-wise architecture in detail.

\subsection{High-level Overview}\label{sec:hlevel}
\noindent \ourm comprises two components -- (i) embedding initiator (EI) and (ii) sequential recommender (SR). Figure~\ref{fig:overall} shows the overall architecture of \ourm with different components and a schematic diagram for both is given in Figure~\ref{fig:model}. The workflow of \ourm includes three steps: (i) determining the embeddings of all POI and app categories using the EI module; (ii) calculating the relative positional encodings in terms of app category, POI category, and time of \cin, and determining embedding matrices for each; and (iii) using the category embeddings from EI module and the newly derived relative embeddings to determine the mobility preferences of a user via the SR module.

As we model the differences between the category of POI and smartphone apps across \cins, we must capture the semantic meaning associated with each category, for \eg\ the difference between a \cat{Sushi restaurant} and a \cat{Cafe}. Accordingly, EI takes the \cin sequence of a user as input, learns the representations of all smartphone app- and POI-categories, and calculates the evolving user preferences as the variation between \cins.
\begin{equation}
\bs{A}, \bs{S} = G_{\mathrm{EI}}(\cm{E}_k, \cm{A}^*_k, \cm{S}^*_k),
\end{equation}
where $\bs{A}, \bs{S}$ denote the learned embeddings for app and POI categories respectively, and $G_{\mathrm{EI}}(\bullet)$ denotes the Embedding Initiator. Moreover, \ourm works by modeling the variations between different \cins in a sequence. Specifically, it learns how the mobility preference of a user has evolved based on the difference in the current and past \cins. Capturing and feeding these differences to our self-attention model is a challenging task as we denote each \cin via POI and app category embeddings. Therefore, we derive these differences and simultaneously embed them to be fed into a self-attention model. These variations are used to assign relative positional encodings to the \cins in the stacked self-attention architecture in SR.
\begin{equation}
\bs{J}, \bs{K}, \bs{T} = f_{\mathrm{RE}}(\cm{E}_k, \bs{A}, \bs{S}),
\end{equation}
where $\bs{J}, \bs{K}, \bs{T}$ are the relative positional encodings for app categories, POI categories, and time respectively. Here, $f_{\mathrm{RE}}(\bullet)$ denotes the function to calculate these relative encodings. Note that these encodings are \textit{personalized}, \ie, they are calculated independently for each user. SR  then combines these relative encodings with absolute positional encodings to model the sequential POI preference of a user. Through this, we aim to get a ranked list of the most probable candidate POIs for the next \cin of a user. 
\begin{equation}
\widehat{l_{k+1}} = G_{\mathrm{SR}}(\cm{E}_k, \bs{J}, \bs{K}, \bs{T}),
\end{equation}
where $\widehat{l_{k+1}}$ is the candidate POI for the $k+1$-th \cin of a user and $G_{\mathrm{SR}}(\bullet)$ denotes the sequential recommender. Figure \ref{fig:model} shows a schematic diagram of \ourm architecture. The training process of \ourm is divided into two steps -- train the category embeddings using EI and then use them for sequential recommendation in the SR section. More details are given in Section \ref{sec:train}.

\xhdr{Preserving the user's privacy via \ourm} Here, we highlight the privacy-conscious nature of our underlying framework. In detail, the existing techniques that model the relationship between the app usage and the physical mobility of a user utilize the precise geo-coordinates, accurate apps used, and the details of all background apps~\cite{reapp, li, f2f}. Thus, these approaches have two major drawbacks: (i) they are a serious violation of the privacy of a user; and (ii) collecting accurate data of this granularity makes the problem highly synthetic in nature. Therefore, in \ourm, we do not incorporate any of this information that can compromise the privacy of a user. Specifically, we only use the category of the visited POI and the category of the only active app. Therefore, in our setting, it is difficult to identify individual users based on these coarse-grained records. Moreover, such cross-app data can be collected while simultaneously preserving the user's privacy~\cite{cross}.

\begin{table}[t!]
\small
\caption{\label{tab:par} Summary of Notations Used.}
\vspace{-2mm}
\centering
\resizebox{0.95\columnwidth}{!}{
\begin{tabular}{ll}
\toprule
\textbf{Notation} & \textbf{Description}\\
\midrule
$\cm{U}, \cm{L}$ & Set of all user and locations\\
$e_k \in \cm{E}$ & $k$-th \cin in the sequence\\
$\cm{A}, \cm{S}$ & Set of all smartphone and POI categories\\
$\bs{A}, \bs{S}$ & Smartphone and POI category embeddings\\
$D$ & Embedding Dimension \\
$\bs{J}, \bs{K}, \bs{T}$ & App, POI, and time based relative encodings\\
$\bs{P}^{\mathrm{key}}, \bs{P}^{\mathrm{val}}$ & Absolute positional encodings\\
$\bs{J}^{\mathrm{key}}, \bs{K}^{\mathrm{key}}, \bs{T}^{\mathrm{key}}$ & Key matrices for relative embeddings\\
$\bs{J}^{\mathrm{val}}, \bs{K}^{\mathrm{val}}, \bs{T}^{\mathrm{val}}$ & Value matrices for relative embeddings\\
$\mathscr{L}_{\mathrm{MF}}, \mathscr{L}_{\mathrm{Bert}}$ & Trajectory and BERT-based loss\\
$\mathscr{L}_{\mathrm{Rec}}$ & POI recommendation loss for SR\\
$\mathscr{L}_{\mathrm{App}}, \mathscr{L}_{\mathrm{POI}}$ & Smartphone and POI category loss\\
\bottomrule
\end{tabular}
}
\vspace{-0.4mm}
\end{table}

\subsection{Embedding Initiator (EI)} \label{sec:eim}
\noindent A major contribution of this paper is to learn the mobility preferences conditioned only on the categories of smartphone apps and POI rather than the exact location coordinates and app preferences. Learning from such coarse data is not a trivial task and training with \textit{randomly-initialized} embeddings may not capture the category semantics, \eg\ if \cat{Burger-Joints} and \cat{Asian-Restaurants} are frequently visited then a training process with random initialization will lead to similar the trained embeddings. Therefore, our category embeddings must simultaneously capture the user preferences towards each category and the category semantics via pre-trained word embeddings. We highlight this through an example -- a user checks a mobile-app of category \cat{Social} frequently at two separate locations, say \cat{Cafe} and \cat{Sushi Restaurant}, the category embeddings should capture the POI influence that persuaded a user to use apps of a similar category(\cat{Social} in this case) as well as the semantic difference between a coffee joint and an Asian restaurant. Therefore, we use a two-channel training procedure, wherein we use pre-trained embeddings to extract the semantic meaning of all app and location categories and learn user preferences towards these categories via a lightweight matrix factorization. Specifically, given a \cin sequence $e_k \in \cm{E}$ we follow a four-layer architecture:

\begin{asparaenum}[(1)]
\item \textbf{Input Layer.} We initially embed the app and location categories, $\cm{A}^*$ and $\cm{S}^*$, as $\bs{A} \in \mathbb{R}^{|\cm{A}^*| \times D}$ and $\bs{S} \in \mathbb{R}^{|\cm{S}^*| \times D}$ respectively. Each row $\bs{a}_i \in \bs{A}$ represents a $D$-dimension representation of a smartphone app category. Similarly, $\bs{s}_i \in \bs{S}$ is a representation for a POI category.

\item \textbf{MF Layer.} To learn the interaction between the app and POI categories, we follow a lightweight collaborative filtering approach, wherein we concatenate the entries in $\bs{A}$ and $\bs{S}$ that appear together in a \cin $e_k \in \cm{E}$. Specifically, we concatenate the app and POI category embeddings for a \cin and then use a feed-forward network. 
\begin{equation}
\widehat{v}_{a_i, s_j} = \mathrm{ReLU} \left( \bs{w}_{v}(\bs{a}_i || \bs{s}_j) + \bs{b}_v \right), 
\end{equation}
where $\widehat{v}_{a_i, s_j}$  denotes the probability of an app of category $a_i$ to be accessed at a POI of category $s_j$, $||$ denotes the concatenation operator, and $\bs{w}_{\bullet}, \bs{b}_{\bullet}$ are trainable parameters. We train our embeddings via a cross-entropy loss:
\begin{multline}
\mathscr{L}_{\mathrm{MF}} = -\sum_{k=1}^{|\cm{E}|} \sum_{\substack{a_i \in \cm{A}_k \\ s_j \in \cm{S}_k}} \bigg [ \log \big( \sigma(\widehat{v}_{a_i, s_j}) \big) + \log \big( 1 - \sigma(\widehat{v}_{a_i, s'_j}) \big) \\ \hspace{4cm}
+ \log \big( 1 - \sigma(\widehat{v}_{a'_i, s_j}) \big) \bigg ],
\end{multline}
where $\widehat{v}_{\bullet, \bullet}$ denotes the estimated access probability (i) $\widehat{v}_{a_i, s_j}$ between a \textit{true} app- and location-category, \ie, $a_i \in \cm{A}_k, s_j \in \cm{S}_k$, (ii) $\widehat{v}_{a'_i, s_j}$ for a negatively sampled app-category with a true location-category, \ie, $a_i \notin \cm{A}_k, s_j \in \cm{S}_k$, and (iii) $\widehat{v}_{a_i, s'_i}$ for a negatively sampled location-category with a true app-category, \ie, $a_i \in \cm{A}_k, s_j \notin \cm{S}_k$.

\item \textbf{BERT Layer.}
To capture the real-world semantics of a category, we use a pre-trained BERT~\cite{bert} model with over 110M parameters. Specifically, we extract the embeddings for each smartphone app and POI category from the pre-trained model. Later, we maximize the similarity between these embeddings and our category representations, $\bs{A}$ and $\bs{S}$ by optimizing a mean squared loss.
\begin{equation}
\mathscr{L}_{\mathrm{Bert}} = \frac{1}{|\cm{E}|} \sum_{k=1}^{|\cm{E}|} \sum_{\substack{a_i \in \cm{A}_k \\ s_j \in \cm{S}_k}} \left [||\bs{a}_i - \Phi_1(a_i)||^2 + || \bs{s}_i - \Phi_2(s_i)||^2 \right],
\end{equation}
where, $\bs{a}_i \in \bs{A}$ and $\bs{s}_j \in \bs{S}$ are our trainable embedding for categories $a_i$ and $s_j$ respectively, and $\Phi_{\bullet}$ denotes a two-step function that extracts pre-trained embeddings for all categories and uses a feed-forward network to normalize the embedding dimension to $D$. Specifically,
\begin{equation}
\Phi_1(a_i) = \mathrm{ReLU} \big(\bs{w}_1 \cdot \cm{B}(a_i) + \bs{b}_1 \big),
\end{equation}
\begin{equation}
\Phi_2(s_i) = \mathrm{ReLU} \big(\bs{w}_2 \cdot \cm{B}(s_i) + \bs{b}_2\big),
\end{equation}
where $\cm{B}$ denotes the set of all pre-trained embeddings, $\cm{B}(a_i)$ and $\cm{B}(s_i)$ denote the extracted app and location category embedding, and $\bs{w}_{\bullet}, \bs{b}_{\bullet}$ are trainable parameters.

\item \textbf{Optimization.}
We train our embeddings using a two-channel learning procedure consisting of app-location interaction loss, $\mathscr{L}_{\mathrm{MF}}$, and pre-trained embedding loss, $\mathscr{L}_{\mathrm{Bert}}$, by optimizing a \textit{weighted} joint loss.
\begin{equation}
\mathscr{L}_{\mathrm{EI}} \,=\, \gamma\mathscr{L}_{\mathrm{MF}} + (1 - \gamma) \mathscr{L}_{\mathrm{Bert}},
\end{equation}
where $\gamma$ denotes a scaling parameter. Later, we use $\bs{A}$ and $\bs{S}$ to identify the inter-\cin differences and model the POI preferences of a user.
\end{asparaenum}

\subsection{Relative or Inter-\cin Variations}
\noindent Buoyed by the efficacy of relative encodings for self-attention models~\cite{postran,reltran}, \ourm captures the evolving preferences of a user as relative encodings based on three inter-\cin differences: (i) Smartphone App-based dynamics, (ii) Location category distribution, and (iii) Time-based evolution across the event sequence.

\xhdr{Smartphone App-based Variation}
Recent research~\cite{postran,reltran} has shown that users' preferences towards smartphone apps are influenced by their geo-locations and other POI-based semantics. Seemingly, it is more likely for a user to be active on a multiplayer game at a social joint rather than at her workplace. We quantify the differences in the app preferences of a user via the differences in the embeddings of the smartphone-app category being used at a \cin. However, in our datasets, every smartphone app is associated with at least one category, \ie, an app can belong to multiple categories, for \eg, Amazon belongs to only one category of \cat{Retail}, but PUBG (a popular mobile game) may belong to categories \cat{Game} and \cat{Action Game}. Calculating the variation based on different embedding is a challenging task. Therefore, we first calculate a ``net app-category'' embedding to denote the representation of all categories an app belongs to. Specifically, for each \cin $e_k$, we calculate the \textit{net} app-category as a mean of all category embeddings.
\begin{equation}
\bs{\mu}^a_k = \frac{1}{|\cm{A}_k|} \sum_{a_i \in \cm{A}_k}\bs{a}_i,
\end{equation}
where $\bs{\mu}^a_k, a_i \in \cm{A}_k, \bs{a}_i \in \bs{A}$ represent the net app-category embedding for a \cin $e_k$, the app-category used in the \cin and the corresponding embedding learned in the EI (see Section~\ref{sec:eim}). Such an embedding allows us to simplify the input given to the self-attention mechanism in our recommender system. Following~\cite{reltran}, we use these embeddings to calculate a inter-\cin variance matrix $\bs{J} \in \mathbb{W}^{|\cm{E}| \times |\cm{E}|}$ for each \cin sequence. Specifically, the $i$-th row in matrix $\bs{J}$ denotes the difference between the mean app-category embedding of \cin $e_i$ with all other \cins in the sequence and is calculated as:
\begin{equation}
\bs{J}_{i,j} = \Bigg\lfloor \frac{ f_{\mathrm{cos}} (\bs{\mu}^a_i, \bs{\mu}^a_j) - \min_f(\cm{E})}{ \max_f(\cm{E}) - \min_f(\cm{E})} \cdot I_a \Bigg\rfloor,
\label{eq:J}
\end{equation}
where $f_{\mathrm{cos}}(\bullet, \bullet), \min_f(\cm{E}), \max_f(\cm{E})$ denote the function for normalized cosine-distance, the minimum and maximum cosine distance between the mean category embedding for any two \cins in a sequence. We use $I_a$ as a clipping constant and a \textit{floor} operator to discretize the entries in $\bs{J}$. Such discretization makes it convenient to extract positional encodings for the self-attention model in SR. 

\xhdr{POI-based Variation} We derive the inter-\cin differences between POI categories using a similar procedure for app-based differences. However,  POI can belong to multiple categories. Therefore, similar to our procedure for calculating app-based variations, we calculate a \textit{net} POI category embedding, $\bs{\mu}^l_k$ for each \cin as $\bs{\mu}^l_k = \frac{1}{|\cm{S}_k|} \sum_{s_i \in \cm{S}_k}\bs{s}_i$. Later, as in Eqn~\ref{eq:J}, we calculate the POI-based inter-\cin variance matrix $\bs{K} \in \mathbb{W}^{|\cm{E}| \times |\cm{E}|}$ using a clipping constant $I_l$. Here, the $i$-th row in matrix $\bs{K}$ denotes the difference between the mean POI-category embedding of \cin $e_i$ with all other \cins in the sequence.

\xhdr{Time-based Variation} Ostensibly, there may be irregularities in the smartphone app usage of a user, \eg\ a user browsing \cat{Amazon} may receive a message \cat{Twitter} that she immediately checks and then later continues her shopping on Amazon. Notably, the \cat{Amazon} app did not influence the user to access \cat{Twitter} and vice-versa, as such a change between apps was coincidental. To model these nuances in \ourm we use the time interval between accessing different smartphone apps. Specifically, similar to app- and POI-category based inter-\cin differences, we derive a \textit{time-based} variations matrix, $\bs{T} \in \mathbb{W}^{|\cm{E}| \times |\cm{E}|}$, using the absolute time-difference between each \cin.
\begin{equation}
\bs{T}_{i,j} = \Bigg\lfloor \frac{|t_i - t_j|}{t_{\min}} \cdot I_t \Bigg\rfloor,
\end{equation}
where $t_i, t_j, t_{\min},$ and $I_t$ denote the time of \cin $e_i$ and $e_j$, minimum time-interval between \cins of a user and the normalizing constant for time respectively. 

\subsection{Sequential Recommender (SR)}
\noindent In this section, we elaborate on the sequential recommendation procedure of \ourm that is responsible for modeling the app and POI preferences of a user and then recommend candidate POI for the next \cin. Specifically, it uses a self-attention architecture consisting of five layers:

\begin{asparaenum}[(1)]
\item \textbf{Input Layer.} The SR model takes the \cin sequence of a user ($\cm{E}$), relative app, POI, and time encodings, ($\bs{K}, \bs{J}$, and $\bs{T}$ respectively), and the \textit{mean} app and location category representations ($\bs{\mu}^a_{\bullet}, \bs{\mu}^l_{\bullet}$) as input to the self-attention model. Since the self-attention models require a fixed input sequence, we limit our training to a fixed number of \cins, \ie, we consider the $N$ most recent \cins in $\cm{E}$ for training our model and if the number of \cins is lesser than $N$, we repeatedly add a \textit{[pad]} vector for the initial \cins within the sequence. 

\item \textbf{Embedding Retrieval Layer.} 
Since the self-attention models are oblivious of the position of each \cin in the sequence, we use a trainable positional embedding for each \cin~\cite{sasrec, tisasrec}. Specifically, we initialize two distinct vectors denoted by $\bs{P}^{\mathrm{key}} \in \mathbb{R}^{N \times D}$ and $\bs{P}^{\mathrm{val}} \in \mathbb{R}^{N \times D}$ where the $i$-th rows, $\bs{p}^{\mathrm{key}}_i$ and $\bs{p}^{\mathrm{val}}_i$, denote the positional encoding for the \cin $e_i$ in the sequence. 
Similarly, we embed the relative positional matrices $\bs{K}$, $\bs{J}$, and $\bs{T}$ into encoding matrices $\bs{J}^{\mathrm{key}}, \bs{J}^{\mathrm{val}} \in \mathbb{R}^{N \times N \times D}$, $\bs{K}^{\mathrm{key}}, \bs{K}^{\mathrm{val}} \in \mathbb{R}^{N \times N \times D}$, and $\bs{T}^{\mathrm{key}}, \bs{T}^{\mathrm{val}} \in \mathbb{R}^{N \times N \times D}$ respectively.
\begin{equation}
\bs{K}^{\mathrm{key}} = \begin{bmatrix} \bs{k}^{\mathrm{key}}_{1,1} & \cdots & \bs{k}^{\mathrm{key}}_{1,N}\\ \vdots & \vdots & \vdots \\ \bs{k}^{\mathrm{key}}_{N,1} & \cdots & \bs{k}^{\mathrm{key}}_{N,N} \end{bmatrix}, \, \bs{K}^{\mathrm{val}} = \begin{bmatrix} \bs{k}^{\mathrm{val}}_{1,1} & \cdots & \bs{k}^{\mathrm{val}}_{1,N}\\ \vdots & \vdots & \vdots \\ \bs{k}^{\mathrm{val}}_{N,1} & \cdots & \bs{k}^{\mathrm{val}}_{N,N} \end{bmatrix},
\label{rel_emb}
\end{equation}
We use two separate matrices to avoid any further linear transformations~\cite{reltran}. Each entry in $\bs{K}^{\mathrm{key}}$ and $\bs{K}^{\mathrm{val}}$ denotes a $D$ dimensional vector representation of corresponding value in  in $\bs{K}$. We follow a similar procedure to initialize $\bs{J}^{\mathrm{key}}, \bs{J}^{\mathrm{val}}, \bs{T}^{\mathrm{key}}$ and $\bs{T}^{\mathrm{val}}$ for $\bs{J}$ and $\bs{T}$ respectively.

\item \textbf{Self-Attention Layer.}
Given the \cin sequence of a user, the self-attention architecture learns the sequential preference of a user towards POIs. Specifically, for an input sequence consisting of POI embeddings of locations visited by a user, $\bs{L}^{\cm{E}} = (\bs{l}_{e_1}, \bs{l}_{e_2}, \cdots \bs{l}_{e_N})$ where $l_{e_i} \in e_i$ and $\bs{l}_{e_i} \in \bs{L}$ are the location visited in \cin $e_i$ the POI embedding for $l_{e_i}$ respectively, we compute a new sequence $\bs{Z} = (\bs{z}_1, \bs{z}_2, \cdots \bs{z}_N)$, where $\bs{z}_{\bullet} \in \mathbb{R}^{D}$. Each output embedding is calculated as a weighted aggregation of embeddings of all POIs visited in the past. 
\begin{equation}
\bs{z}_i = \sum_{j=1}^{N} \bs{\alpha}_{i,j} \left( \bs{w}_{v, j}\bs{l}_{e_j} + \overline{\bs{\mu}}_j + \bs{p}^{\mathrm{val}}_j + \bs{j}^{\mathrm{val}}_{i,j} + \bs{k}^{\mathrm{val}}_{i,j} + \bs{t}^{\mathrm{val}}_{i,j} \right),
\label{eq:tv}
\end{equation}
where $\bs{l}_{e_j}$ is the POI embedding, $\overline{\bs{\mu}}_j = \bs{\mu}^a_j + \bs{\mu}^l_j$ is the sum of smartphone app and POI category mean embeddings, and $\bs{w}_{v, j}$ is a trainable parameter. The attention weights $\bs{\alpha}_{\bullet, \bullet}$ are calculated using a soft-max over other input embeddings as:
\begin{equation}
\bs{\alpha}_{i,j} = \frac{\exp \big(x_{i,j}\big)}{\sum_{k=1}^{N} \exp \big( x_{i,k}\big)},
\label{eq:tsoft}
\end{equation}
where $x_{i,j}$ denotes the compatibility between two \cins -- $e_i$ and $e_j$ -- and is computed using both -- relative- as well as absolute-positional encodings.
\begin{equation}
x_{i,j} = \frac{\bs{w}_{q, i}\bs{l}_{e_i} \left( \bs{w}_{k, j}\bs{l}_{e_j} + \bs{p}^{\mathrm{key}}_j + \bs{j}^{\mathrm{key}}_{i,j} + \bs{k}^{\mathrm{key}}_{i,j} + \bs{t}^{\mathrm{key}}_{i,j} \right )^{\top}}{\sqrt{D}},
\label{eq:tqk}
\end{equation}
where $\bs{w}_{q, \bullet}, \bs{w}_{k, \bullet}$ and $D$ denote the input \textit{query} projection, \textit{key} projection, and the embedding dimension respectively. We use the denominator as a scaling factor to control the dot-product gradients. As our task is to recommend candidate POI for future \cins and should
only consider the first $k$ \cins to predict the $(k + 1)$-th \cin, we introduce a \textit{causality} over the input sequence. Specifically, we modify the attention procedure in Eqn. (\ref{eq:tqk}) and remove all links between the future and the current \cin.

\item \textbf{Point-wise Layer.} As the self-attention lacks any non-linearity, we apply a feed-forward layer with two linear-transformation with ReLU activation. 
\begin{equation}
\mathrm{PFFN}(\bs{z}_k) = \mathrm{ReLU} \left ( \bs{z}_k \bs{w}_{p,1} + \bs{b}_{1} \right ) \bs{w}_{p,2} + \bs{b}_2, 
\end{equation}
where $\bs{w}_{p, \bullet}, \bs{b}_{\bullet}$ are trainable layer parameters.

The combination of a self-attention layer and the point-wise layer is referred to as a self-attention \textit{block} and stacking self-attention blocks gives the model more flexibility to learn complicated dynamics~\cite{transformer}. Thus, we stack $M_b$ such blocks, and to stabilize the learning process, we add a residual connection between each such block.
\begin{equation}
\bs{z}^{(r)}_k = \bs{z}^{(r-1)}_k + \mathrm{PFFN} \big( f_{ln} (\bs{z}^{(r-1)}_k) \big),
\end{equation}
where $1 \le r \le M_b, f_{ln}(\bullet)$ denote the level of the current self-attention block and \textit{layer-normalization} function respectively. The latter is used to further accelerate the training of self-attention and is defined as follows:
\begin{equation}
f_{ln}(\bs{z}_k) = \beta \odot \frac{\bs{z}_k - \mu_z}{\sqrt{\sigma_z^2 + \epsilon}} + \xi,
\end{equation}
where $\odot, \mu_z, \sigma_z, \beta, \xi, \epsilon$ denote the element-wise product, mean of all input embeddings, the variance of all input embeddings, learned scaling factor, bias term, and the Laplace smoothing constant respectively.

\item \textbf{Prediction Layer.} A crucial distinction between \ourm and the standard self-attention model is that we not only predict the candidate POIs for the next \cin, but also the category of the smartphone app and POI category to be used in the next \cin. We describe the prediction procedure as follows:

\xhdr{POI Recommendation} We predict the next POI to be visited by a user in the \cin sequence using a matrix-factorization~\cite{ncf} based approach between the transformer output $\bs{Z}^{(M_b)} = (z_1, z_2, \cdots z_k)$ and the embeddings of POIs visited by the user, $(\bs{l}_{e_2}, \bs{l}_{e_{3}}, \cdots \bs{l}_{e_{k+1}})$
\begin{equation}
\widehat{v_{u_i, l_{e_k}}} = \bs{z}_{k-1} \bs{l}_{e_{k}}^{\top},
\end{equation}
where $\widehat{v_{u_i, l_{e_k}}}$ is the calculated probability of user, $u_i$, to visit the POI, $l_{e_k}$, for her next \cin. We learn the model parameters by minimizing the following cross-entropy loss.
\begin{multline*}
\mathscr{L}_{\mathrm{Rec}} = -\sum_{u_i \in \cm{U}} \sum_{k=1}^{N} \left[ \log \left ( \sigma(\widehat{v_{u_i, l_{e_k}}} \right) + \log \left (1 - \sigma(\widehat{v_{u_i, l'_{e_k}}}) \right) \right] \\ + \lambda || \Theta ||_F^2,
\label{eq:cross}
\end{multline*}
where $\widehat{v_{u_i, l'_{e_k}}}$ denotes the \cin probability for a negatively sampled POI, \ie, a randomly sampled location that will not be visited by a user. $\lambda$ and $\Theta$ denote the regularization parameter and the trainable parameters respectively.

\xhdr{Predicting App Categories} Predicting the next smartphone app to be accessed by a user has numerous applications ranging from smartphone system optimization, resource management in mobile operating systems, and battery optimization~\cite{appusage2vec, carat}. Therefore, to predict the category of the next app to be used, we follow a matrix-factorization approach to calculate the relationship between the user preference embedding, $\bs{z}_k$, and the mean of smartphone app embeddings for the next \cin. 
\begin{equation}
\widehat{q_{u_i, \cm{A}_k}} = \bs{z}_{k-1} {\bs{\mu}^a_{k}}^{\top},
\end{equation}
where $\widehat{q_{u_i, \cm{A}_k}}, \bs{\mu}^a_{k}$ denote the usage probability of apps of categories in $\cm{A}_k$ and the mean embedding for all apps used in \cin $e_k$. Later, we minimize a cross-entropy loss with negatively sampled apps, \ie, apps that were not used by the user, denoted as $\mathscr{L}_{\mathrm{App}}$.

\xhdr{Predicting Location Categories} As in app-category prediction, we calculate the preference towards a POI-category using the mean of the POI category embedding $\bs{\mu}^l_k$ and learn the parameters by optimizing a similar \textit{cross-entropy} loss denoted as $\mathscr{L}_{\mathrm{POI}}$.

The net loss for sequential recommendation is a weighted combination of POI recommendation loss, app-category loss, and location-category loss.
\begin{equation}
\mathscr{L}_{\mathrm{SR}} = \mathscr{L}_{\mathrm{Rec}} + \kappa (\mathscr{L}_{\mathrm{App}}+ \mathscr{L}_{\mathrm{POI}}),
\end{equation}
Here, $\kappa$ is a tunable hyper-parameter for determining the contribution of category prediction losses. All the parameters of \ourm including the weight matrices, relative-position weights, and embeddings are learned using an Adam optimizer\cite{adam}.
\end{asparaenum}

\subsection{\ourm: Training} \label{sec:train}
\noindent As mentioned in Section~\ref{sec:hlevel}, \ourm involves a two-step training procedure. Specifically, it consists of the following steps: \begin{inparaenum}[(i)] \item learning the app and POI category embeddings using the embedding initiator(EI) and \item training the self-attention model in SR to recommend candidate POI to the user\end{inparaenum}. In detail, we first train the parameters of EI by minimizing the $\mathscr{L}_{EI}$ loss for multiple epochs and later use the trained category embeddings in SR and recommend candidate POI by minimizing the recommendation loss, $\mathscr{L}_{SR}$.

We highlight that a \textit{joint} training of both, EI and SR, is not suitable in the presence of relative positional encodings, as they are conditioned on the category embeddings learned in EI. Therefore, during joint training, an update in the category embedding will make the trained parameters of SR across the previous epochs unsuitable for prediction in the future. Moreover, these encodings are calculated \textit{relatively}, \ie, conditioned on the embedding of other categories in a sequence, and thus any change in the category embedding will affect the category embeddings. 

\section{Experiments}\label{sec:expts}
\noindent In this section, we report a comprehensive empirical evaluation of \ourm and compare it with other state-of-the-art approaches. We evaluate the POI recommendation performance of \ourm using two real-world datasets from China. These datasets vary significantly in terms of data sparsity, the no. of app categories, and POI categories. With our experiments, we aim to answer the following research questions:
\begin{itemize}
\item[\textbf{RQ1}] How does \ourm fare against state-of-the-art models for sequential POI recommendation? Where are the gains and losses?
\item[\textbf{RQ2}] What is the contribution of relative positional encodings?
\item[\textbf{RQ3}] What is the scalability of \ourm and the stability of the learning procedure?
\item[\textbf{RQ4}] What is the impact of different hyper-parameter values on the prediction performance of \ourm?
\end{itemize}
All our algorithms were implemented in Tensorflow~\footnote{\url{www.tensorflow.org} (Accessed July 2022)} and executed on a server running Ubuntu 16.04 with 125GB memory and NVIDIA Tesla V100 32GB GPU.

\subsection{Experimental Setup}
\begin{figure}[t!]
\centering
\vspace{-2mm}
\subfloat[\tel]
{\includegraphics[height=3.2cm]{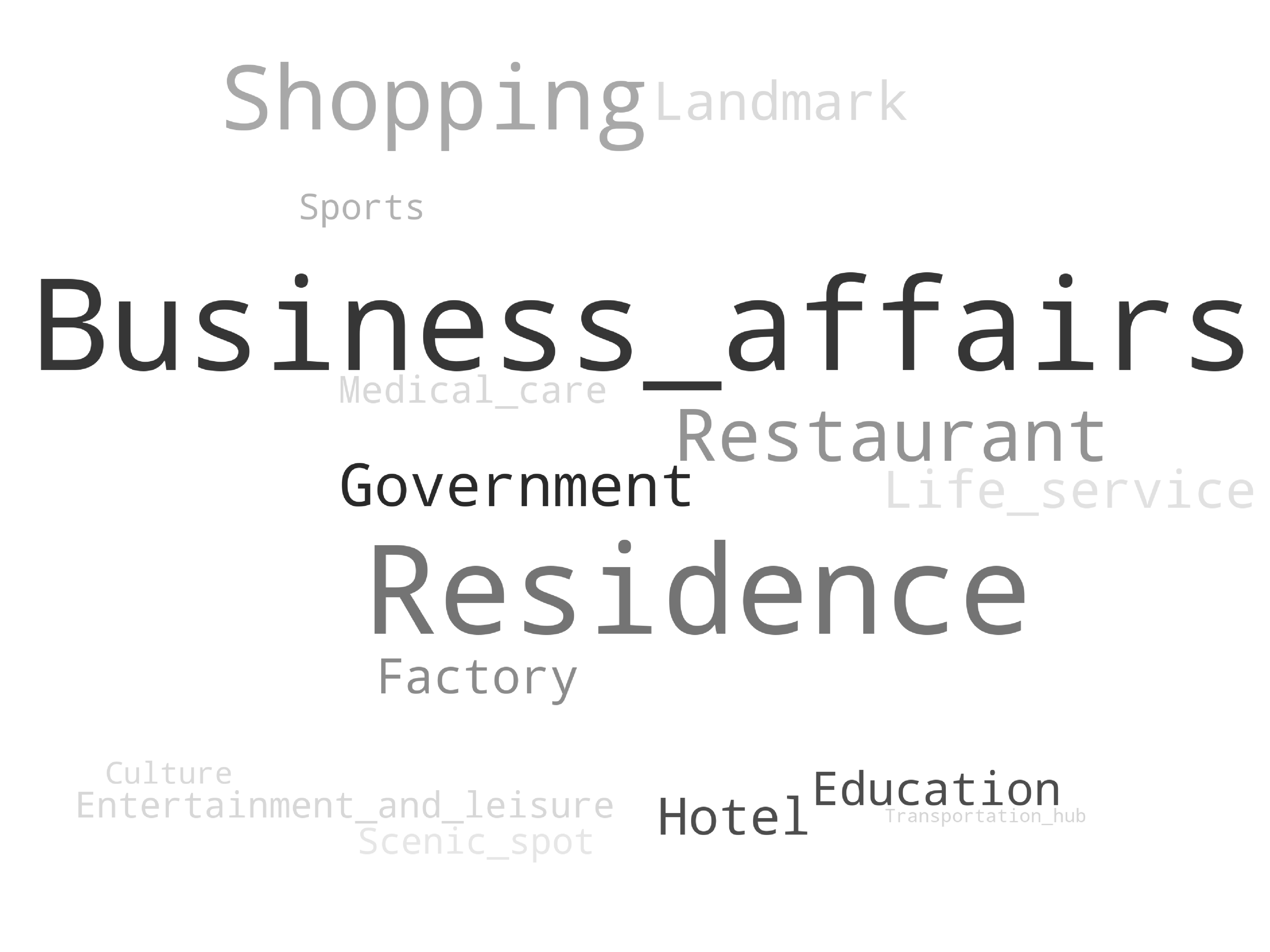}}
\hfill
\subfloat[\tdk]
{\includegraphics[height=3.2cm]{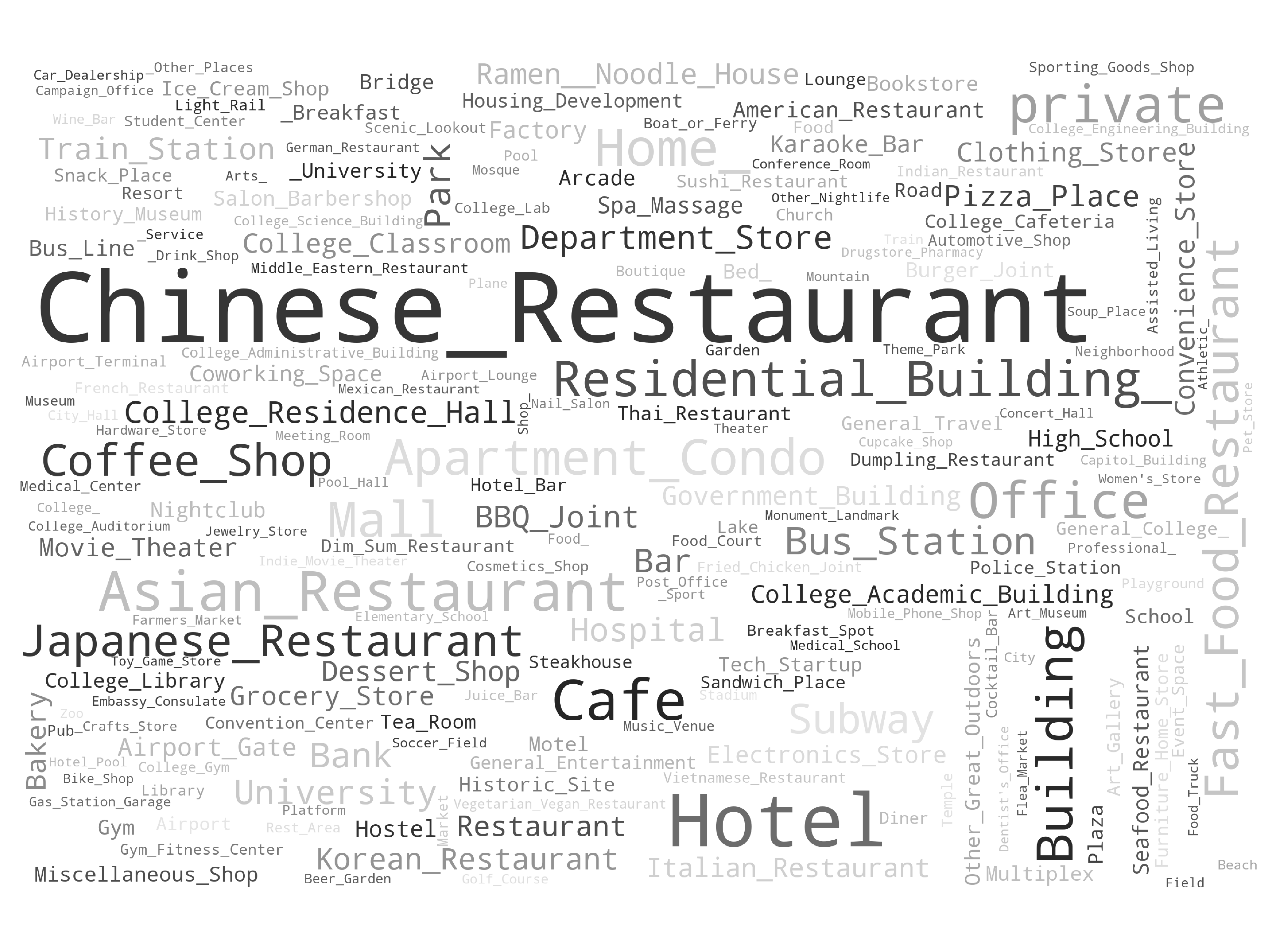}}
\vspace{-2mm}
\caption{POI category-wise word clouds for all datasets. The variance is due to the different information sources, at-hand for \tel and Foursquare for \tdk.}
\vspace{-2mm}
\label{fig:wordcloud}
\end{figure}

\xhdr{Dataset Description}
As our goal is to recommend POIs to a user based on her smartphone usage, the mobility datasets used in our experiments must contain the user trajectory data, \ie, geographical coordinates, time of a \cin, as well as the smartphone-usage statistics -- applications used across different locations, the categories of different apps based on online app-stores, \etc\ Therefore we consider two popular large-scale datasets -- \textit{\tel} and \textit{\tdk} and their statistics are given in Table~\ref{tab:data}. Moreover, we plot the POI-category word clouds in Figure~\ref{fig:wordcloud}.

\begin{asparaenum}[(1)]
\item \textbf{\tel:} This smartphone usage and the physical-mobility dataset was collected by a major network operator in China~\cite{li}. The trajectories were collected from Shanghai in April 2016. It contains the details of a user's physical mobility and the time- and geo-stamped smartphone app usage records. More specifically for each user, we have the time-stamped records of the smartphone apps being used and the different cellular-network base stations to which the smartphone was connected during the data collection procedure. For the region covered by each cellular network base station, we also have the details of the internal POIs and their corresponding categories. For our experiments, we consider each \textit{user$\rightarrow$base-station} entry as a \cin and all the apps and their categories associated with that \cin as the events in the sequence $\cm{E}$. We adopt a commonly followed data cleaning procedure~\cite{axolotl,reformd} and filter out users and POI with less the five \cins.

\item \textbf{\tdk:} A large-scale public app-usage dataset that was released by TalkingData\footnote{\url{www.talkingdata.com/} (Accessed July 2022)}, a leading data intelligence solution provider based in China. The original dataset consists of the location- and time-stamped records of smartphone app usage and physical trajectories of a user~\cite{tk}. However, in this dataset, we lack the categories associated with each POIs. We overcome this by extracting location categories and geo-coordinates from publicly available \cin records~\cite{lbsn2vec} for users in Foursquare -- a leading social mobility network. Specifically, we map every \cin location in Foursquare to a location in the TalkingData within a distance of 50m based on geographical coordinates. For our experiments using this dataset, we restrict our \cin records to only the locations situated in mainland China. As in the \tel dataset, we filter out the users and POI with lesser than five \cins. 
\end{asparaenum}

\begin{table}[t]
\small
\caption{Statistics of all datasets used in this paper. Here, $|\cm{U}|$, $|\cm{P}|$, $|\cm{E}|$, $|\cm{A}|$, and $|\cm{S}|$ denote the number of users, POIs, \cins, app categories, and POI categories respectively.}
\vspace{-2mm}
\begin{tabular}{c|ccccc}
\toprule
\textbf{Dataset} & $|\cm{U}|$ & $|\cm{P}|$ & $|\cm{E}|$ & $|\cm{A}|$ & $|\cm{S}|$ \\ \hline
\tel & 869 & 32680 & 3668184 & 20 & 17 \\
\tdk & 14544 & 37113 & 438570 & 30 & 366 \\ \bottomrule
\end{tabular}
\label{tab:data}
\vspace{-2mm}
\end{table}

\xhdr{Evaluation Metric}
We evaluate \ourm and the other sequential recommendation baselines, using a widely used \textit{leave-one-out} evaluation, \ie, next \cin prediction task. Specifically, for each user, we consider the last \cin of the trajectory sequence as the test \cin, the second last \cin for validation, and all preceding events as the training set~\cite{sasrec, tisasrec}. Since, across both datasets, we have a large number of candidate POIs that a user can visit next. Thus, due to resource constraints and to favor scalability, it is impossible to get a ranked list of every POI corresponding to every user. Therefore, we follow a common testing strategy wherein we pair each ground truth \cin in the test set with $100$ randomly sampled negative events~\cite{ncf, sasrec}. Therefore, the task becomes to rank the negative \cins with the ground truth \cin. We highlight that this is a highly effective approach and consider an analysis over other possible sampling metrics as a future work~\cite{sampled}. In our setting, the hit-rate, HR@$k$, is equivalent to Recall@$k$ and proportional to Precision@$k$, and mean reciprocal rank (MRR) is equivalent to mean average precision (MAP). To evaluate the effectiveness of all approaches, we use Hits@$k$ and NDCG@$k$, with $k \in \{1, 5, 10\}$, and report the confidence intervals based on five independent runs.

\xhdr{Parameter Settings}
For all results in Section~\ref{rq1} and~\ref{rq2}, we set $N = 200$ and $N=100$ for \tel and \tdk respectively. We set $I_a = I_l = I_t = 64$, $D=64$, and $\lambda=0.002$, We search the batch-size in $\{128, 256\}$, the no of attention-heads in $\{1, 2, 4, 8\}$, $\kappa, \gamma $ are searched in $\{ 0.2, 0.5, 0.8\}$, and the dropout probability is set to $0.2$. However, for parameter sensitivity experiments in Section~\ref{rq3}, we show the prediction performance across different hyperparameter values.

\begin{table*}[t]
\centering
\caption{\label{tab:tel_results} Next \cin recommendation performance of \ourm and state-of-the-art baselines. As the \tel dataset lacks precise geographical coordinates for every \cin, we exclude a comparison with STGN~\cite{attnloc4}. Numbers with bold font and superscript * indicate the best and the second best performer respectively. All results of \ourm are statistically significant (i.e. two-sided Fisher's test with $p \le 0.1$) over the best baseline.}
\vspace{-3mm}
\resizebox{0.9\linewidth}{!}{\tabcolsep 3pt
\begin{tabular}{c|ccccc|ccccc}
\toprule
& \multicolumn{5}{c|}{\textbf{\tel}} & \multicolumn{5}{c}{\textbf{\tdk}} \\ \hline 
\textbf{Baselines} & \textbf{NDCG/Hits@1} & \textbf{NDCG@5} & \textbf{NDCG@10} & \textbf{Hits@5} & \textbf{Hits@10} & \textbf{NDCG/Hits@1} & \textbf{NDCG@5} & \textbf{NDCG@10} & \textbf{Hits@5} & \textbf{Hits@10}\\ \hline
FPMC~\cite{fpmc} & 0.5906 & 0.6021 & 0.6402 & 0.6162 & 0.6481 & 0.7224 & 0.7362 & 0.7704 & 0.7408 & 0.7892 \\[0.05cm]
TransRec~\cite{transrec} & 0.5437 & 0.5803 & 0.6055 & 0.5839 & 0.6081 & 0.6872 & 0.6892 & 0.7691 & 0.6902 & 0.7784 \\[0.05cm]
GRU4Rec+~\cite{gru4recplus} & 0.6291 & 0.6432 & 0.6796 & 0.6443 & 0.6867 & 0.7319 & 0.7654 & 0.7913 & 0.7703 & 0.7962 \\[0.05cm]
Caser~\cite{caser} & 0.6418 & 0.6472 & 0.6782 & 0.6507 & 0.6991 & 0.7321 & 0.7802 & 0.8079 & 0.8157 & 0.8482 \\[0.05cm]
STGN~~\cite{attnloc4} & - & - & - & - & - & 0.6694 & 0.7981 & 0.8132 & 0.8224 & 0.8549 \\[0.05cm]
AUM~\cite{reapp2} & 0.5718 & 0.6089 & 0.6358 & 0.6097 & 0.6433 & 0.7184 & 0.7395 & 0.7646 & 0.7782 & 0.8179 \\[0.05cm]
Bert4Rec~\cite{bert4rec} & 0.7031 & 0.7346 & 0.7442 & 0.7188 & 0.7301 & 0.7728 & 0.8247 & 0.8281 & 0.8614 & 0.8743 \\[0.05cm]
SASRec~\cite{sasrec} & 0.7279 & 0.7530 & 0.7562* & 0.7583 & 0.7648 & 0.8295 & 0.8621* & 0.8680 & 0.9027* & 0.9108* \\[0.05cm]
TiSRec~\cite{tisasrec} & 0.7284* & 0.7542* & 0.7558 & 0.7618* & 0.7663* & 0.8307* & 0.8619 & 0.8693* & 0.8998 & 0.9014 \\[0.05cm] \hline
\ourm & \textbf{0.7865} & \textbf{0.8021} & \textbf{0.8186} & \textbf{0.8203} & \textbf{0.8340} & \textbf{0.8793} & \textbf{0.9324} & \textbf{0.9371} & \textbf{0.9492} & \textbf{0.9594} \\ \bottomrule
\end{tabular}
}
\end{table*}

\xhdr{Baselines} 
We compare \ourm with the state-of-the-art methods based on their architectures below:
\begin{compactenum}[(1)]
\item{\bf Standard Recommendation Systems.}
\begin{asparadesc}
\item [FPMC~\cite{fpmc}] FPMC utilizes a combination of factorized first-order Markov chains and matrix factorization for recommendation and encapsulates a user's evolving long-term preferences as well as the short-term purchase-to-purchase transitions.

\item [TransRec~\cite{transrec}] A first-order sequential recommendation model that captures the evolving item-to-item preferences of a user through a translation vector.
\end{asparadesc}

\item{\bf POI Recommendation Systems.}
\begin{asparadesc}
\item [STGN~\cite{attnloc4}] Uses a modified LSTM network that captures the spatial and temporal dynamic user preferences between successive \cins using spatio-temporal gates. Hence, it requires the exact location coordinates as input to the model.
\end{asparadesc}

\item{\bf Smartphone App-based.}
\begin{asparadesc}
\item [AUM~\cite{reapp2}] Models the user mobility as well as app-usage dynamics using a Dirichlet process to predict the next successive \cin locations.
\end{asparadesc}

\item{\bf Recurrent and Convolutional Neural Network.}
\begin{asparadesc}
\item [GRU4Rec+~\cite{gru4recplus}] A RNN-based approach that models the user action sequences for a session-based recommendation. It is an improved version of GRU4Rec~\cite{gru4rec} with changes in the loss function and the sampling techniques.
\item [Caser~\cite{caser}] A state-of-the-art CNN-based sequential recommendation method that applies convolution operations on the $N$-most recent item embeddings to capture the higher-order Markov chains.
\end{asparadesc}

\item{\bf Self-Attention.}
\begin{asparadesc}
\item [Bert4Rec~\cite{bert4rec}] A bi-directional self-attention~\cite{bert} based sequential recommendation model that learn user preferences using a Cloze-task loss function, i.e. predicts the artificially \textit{masks} events form a sequence.
\item [SASRec~\cite{sasrec}] A self-Attention~\cite{transformer} based sequential recommendation method that attentively captures the contribution of each product towards a user's item-preference embedding.
\item [TiSRec~\cite{tisasrec}] A recently proposed enhanced version of the SASRec model that uses \textit{relative}-position embeddings using the difference in the time of consecutive purchases made by the user.
\end{asparadesc}
\end{compactenum}
We omit comparisons with models such as GRU4Rec~\cite{gru4rec} and MARank~\cite{marank} as they already have been outperformed by the current baselines. We calculate the confidence intervals based on the results in three independent runs.

\subsection{Performance Comparison (RQ1)} \label{rq1}
\noindent In Table~\ref{tab:tel_results}, we report the location recommendation performance of different methods across both mobility datasets. From these results, we make the following observations.

\begin{compactitem}[$\bullet$]
\item \ourm consistently outperforms all other baselines for sequential mobility prediction across both datasets. The superior performance signifies the importance of including the smartphone usage pattern of a user to determine her mobility preferences. We also note that the performance gains over other self-attention-based models -- Bert4Rec~\cite{bert4rec}, SASRec~\cite{sasrec}, and TiSRec~\cite{tisasrec} further reinforce our claim that including \textit{relative} positional encodings based on the smartphone, spatial and temporal characteristics enhances the recommendation performance. 

\item We also note that the self-attention-based architecture such as Bert4Rec, SASRec, TiSRec, and \ourm consistently yield the best performance on all the datasets and easily outperform CNN and RNN based models namely Caser~\cite{caser} and GRU4Rec+~\cite{gru4recplus}. This further signifies the unequaled proficiency of the transformer~\cite{transformer} architecture to capture the evolution of user preferences across her trajectory sequence. More importantly, it outperforms the state-of-the-art location recommendation model STGN~\cite{attnloc4} that uses the additional information of precise geographical coordinates of each POI location.

\item \ourm outperforms other smartphone-activity-based approaches AUM~\cite{reapp2} by up to ~34\% across all metrics. 

\item We also note that neural baselines such as Caser~\cite{caser}, GRU4Rec+~\cite{gru4recplus} achieve better results as compared to FPMC~\cite{fpmc} and TransRec~\cite{transrec}. It asserts the utmost importance of designing modern recommender systems using neural architectures. Moreover, GRU4Rec+ achieves a similar performance compared to Caser.
\end{compactitem}

\noindent To sum up, our empirical analysis suggests the following: \begin{inparaenum}[(i)] \item the state-of-the-art models, including self-attention and standard neural models, are not suitable for modeling mobile-user trajectories, and \item \ourm achieves better recommendation performances as it captures the mobility dynamics as well as the smartphone-activity of a user. \end{inparaenum}

\subsection{Ablation Study(RQ2)} \label{rq3}
\noindent We also perform an ablation study to estimate the efficacy of different components in the \ourm architecture. More specifically we aim to calculate the contribution of (i) the embedding initiator and (ii) relative positional embeddings.

\begin{figure}[t]
\centering
\subfloat[\tel]
{\includegraphics[height=2.8cm]{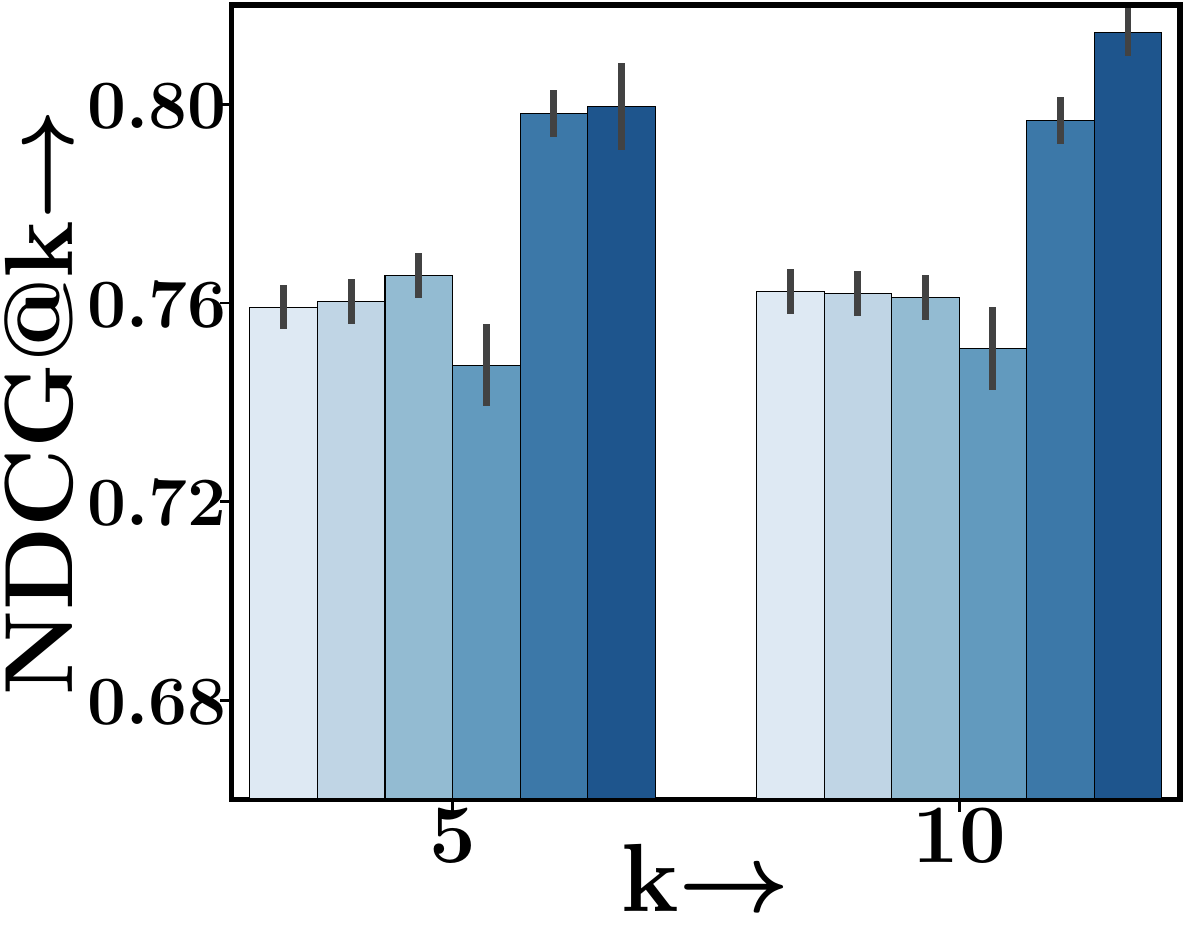}}
\hspace{0.5cm}
\subfloat[\tdk]
{\includegraphics[height=2.8cm]{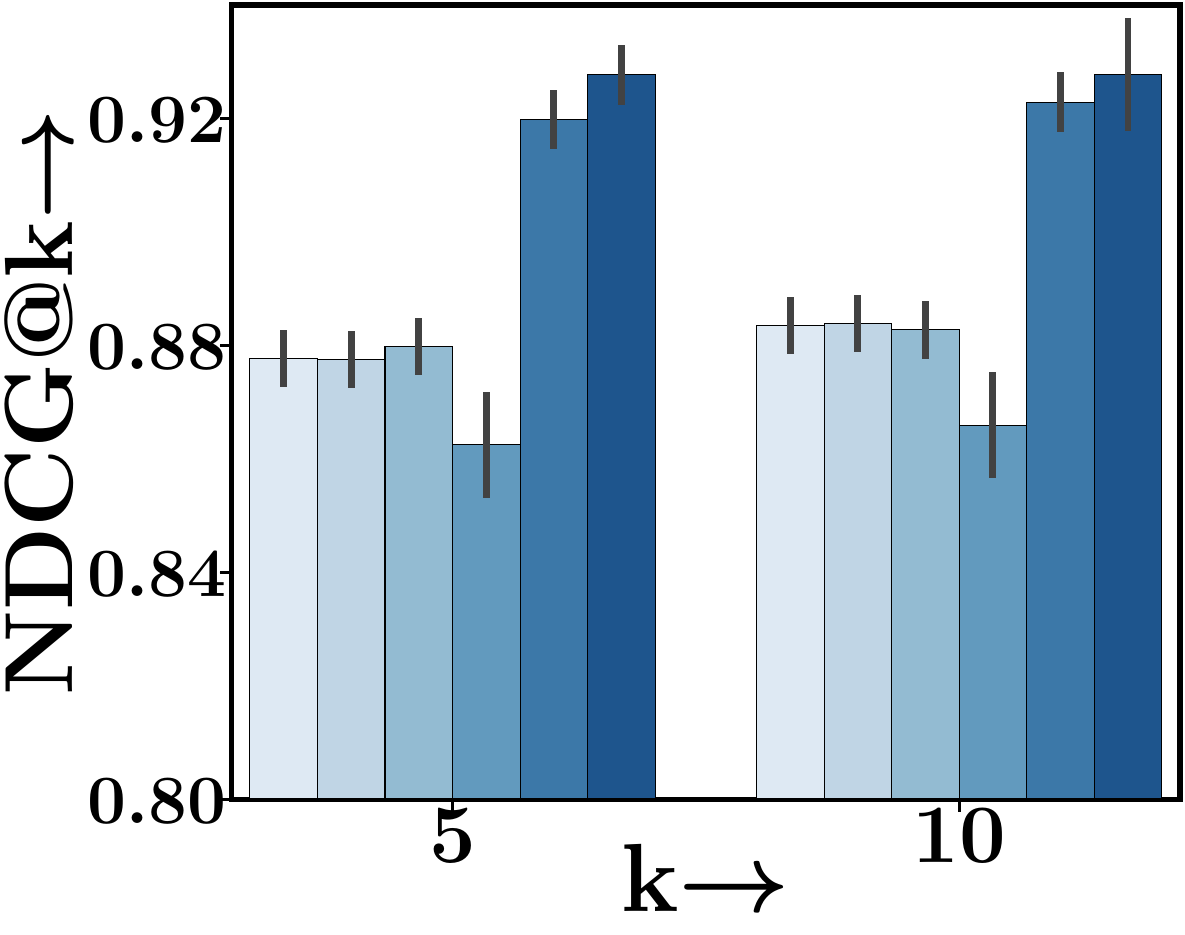}}

{\includegraphics[width=0.8\linewidth]{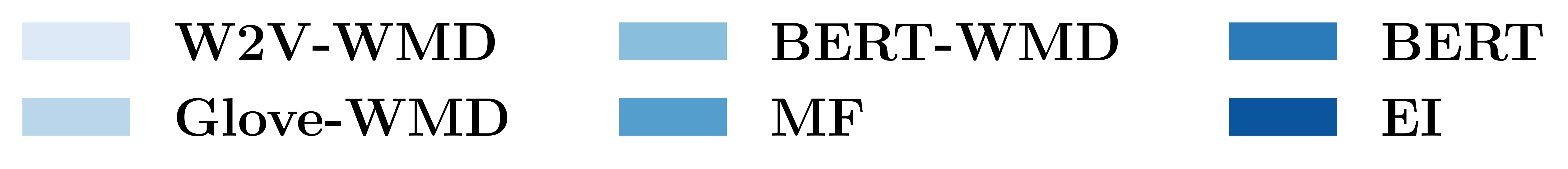}}
\vspace{-2mm}
\caption{POI recommendation performance of \ourm with different methods for obtaining the relative positional encodings, \ie, the inter-\cin differences between app- and location embeddings. Here, the time-based representations are kept consistent across all the models.}
\vspace{-2mm}
\label{fig:init}
\end{figure}

\xhdr{Analysis of Embedding Initiator}
We reiterate that EI, defined in Section \ref{sec:eim}, is used to learn the semantic meaning of each category for POI and apps, as well as the relationship between them in a mobility sequence. To emphasize its importance, we compare the prediction performances of \ourm with different procedures to learn category embeddings and thus the relative embeddings. Specifically we consider: \begin{inparaenum}[(i)] \item word-movers-distance (WMD)~\cite{wmd} between the word2vec~\cite{w2v} representations of each category, \item WMD on Glove~\cite{glove} based representations, \item WMD based on BERT~\cite{bert} initialized vectors, \item a simple collaborative filtering based parameter training, \item using pre-trained BERT, and (vi) the EI proposed in the paper\end{inparaenum}. From the results in Figure~\ref{fig:init}, we note that our proposed EI achieves the best prediction performance compared to other approaches. We also note that standard pre-trained BERT vectors outperform other WMD-based approaches.

\xhdr{Relative Positional Encodings}
Relative positional embeddings are a crucial element in our model. We calculate the performance gains due to the different relative encodings -- app-, time- and  location-based by estimating the recommendation performance of the following approaches: \begin{inparaenum}[(i)] SASRec~\cite{sasrec}; \item TiSRec~\cite{tisasrec}; \item  \ourm with time-based relative positional encoding called \ourm-t; \item  \ourm with app-based encodings, denoted as \ourm-a; \item  \ourm with location-based encodings, denoted as \ourm-l; and \item the complete \ourm model with all relative encodings\end{inparaenum}.

\begin{figure}[t]
\centering
\subfloat[\tel]
{\includegraphics[height=2.8cm]{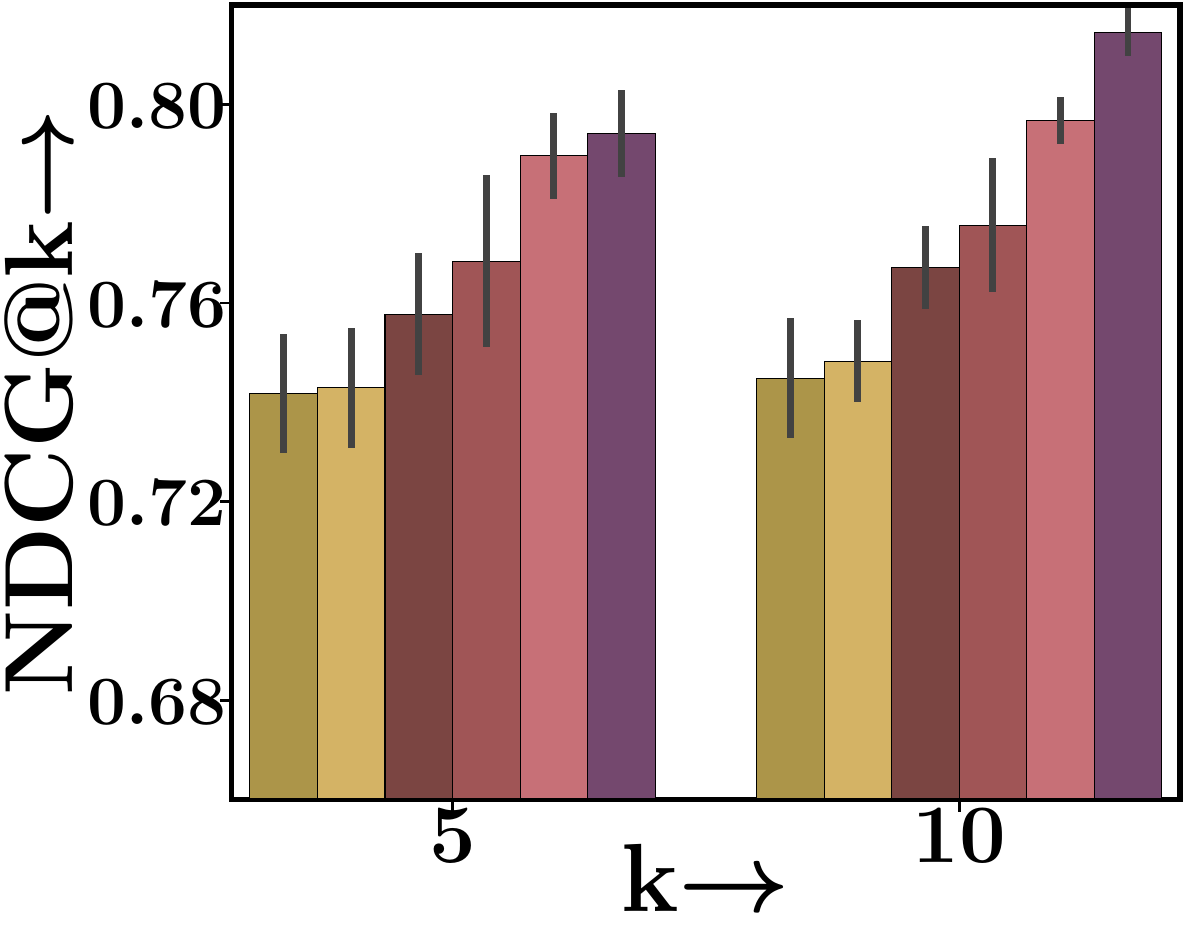}}
\hspace{0.5cm}
\subfloat[\tdk]
{\includegraphics[height=2.8cm]{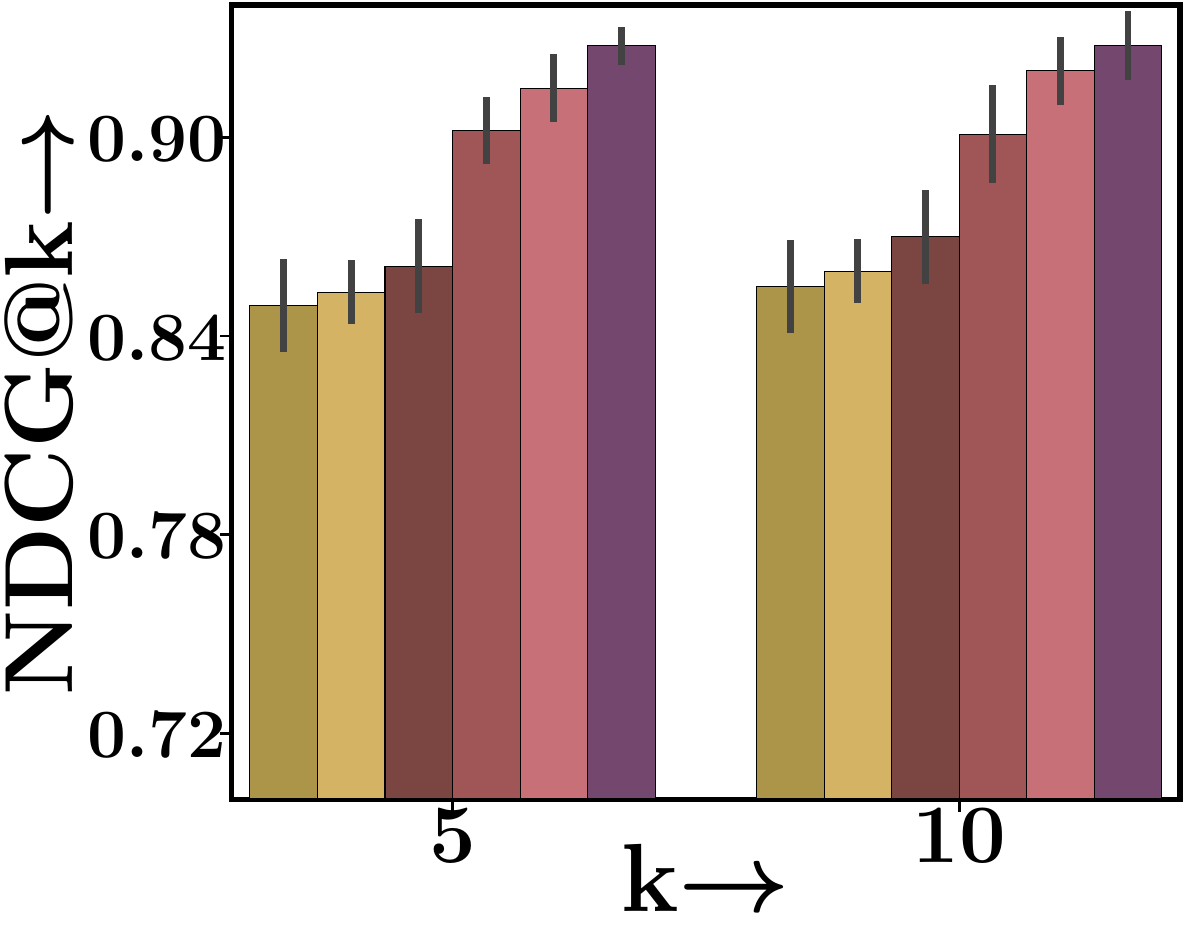}}

{\includegraphics[width=0.8\linewidth]{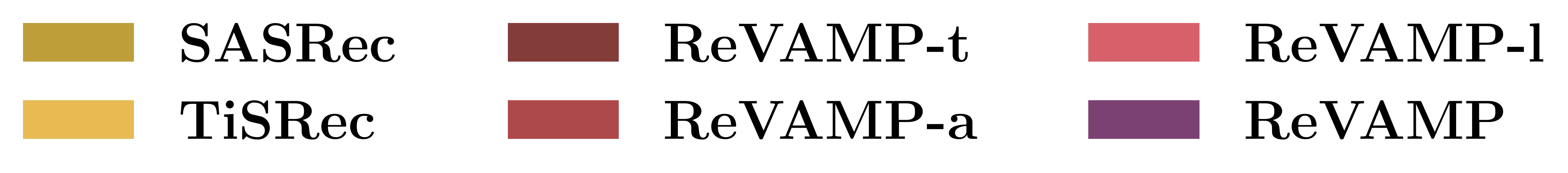}}
\vspace{-2mm}
\caption{Ablation study with different \textit{relative} positional encodings used in \ourm and their comparison with SASRec~\cite{sasrec} and TiSRec~\cite{tisasrec}.}
\vspace{-4mm}
\label{fig:ablation}
\end{figure}
Figure \ref{fig:ablation} summarizes our results in which we observe that including relative positional encodings of any form, whether app-based or location-based, leads to better prediction performances. Interestingly, the contribution of location-based relative positional embeddings is more significant than the app-based and could be attributed to \textit{larger} variations in location-category than the app-category across an event sequence. For example, the difference between location categories of a university region and an office space will effectively capture the larger dynamics than the differences in smartphone app usage across these two regions. However, jointly learning all positional encoding leads to the best performance over both datasets. The improvements of \ourm-t over TiSRec~\cite{tisasrec} could be attributed to the use of \textit{absolute} event encodings (both app and location).

\begin{figure}[t]
\centering
\centering
\subfloat[\tel]
{\includegraphics[height=2.8cm]{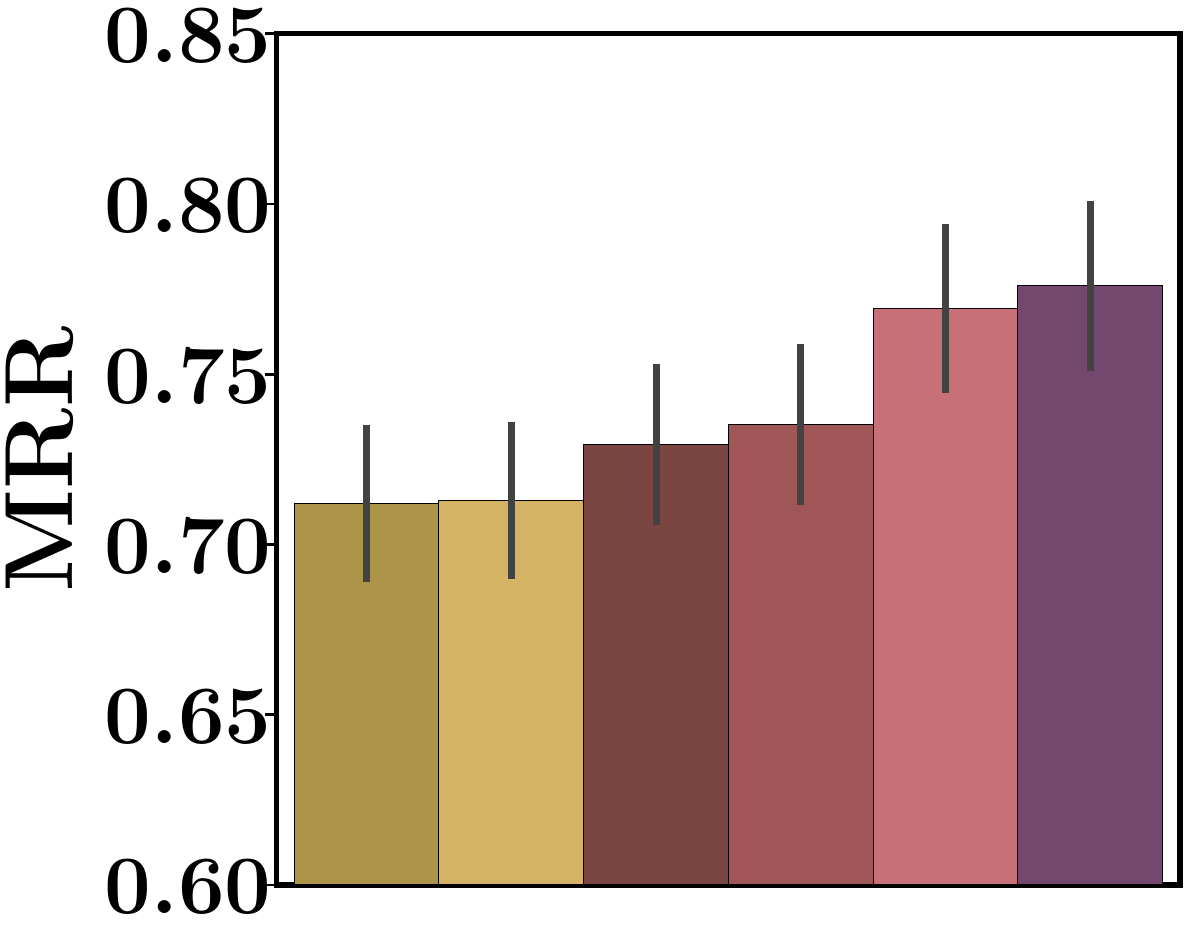}}
\hspace{0.5cm}
\subfloat[\tdk]
{\includegraphics[height=2.8cm]{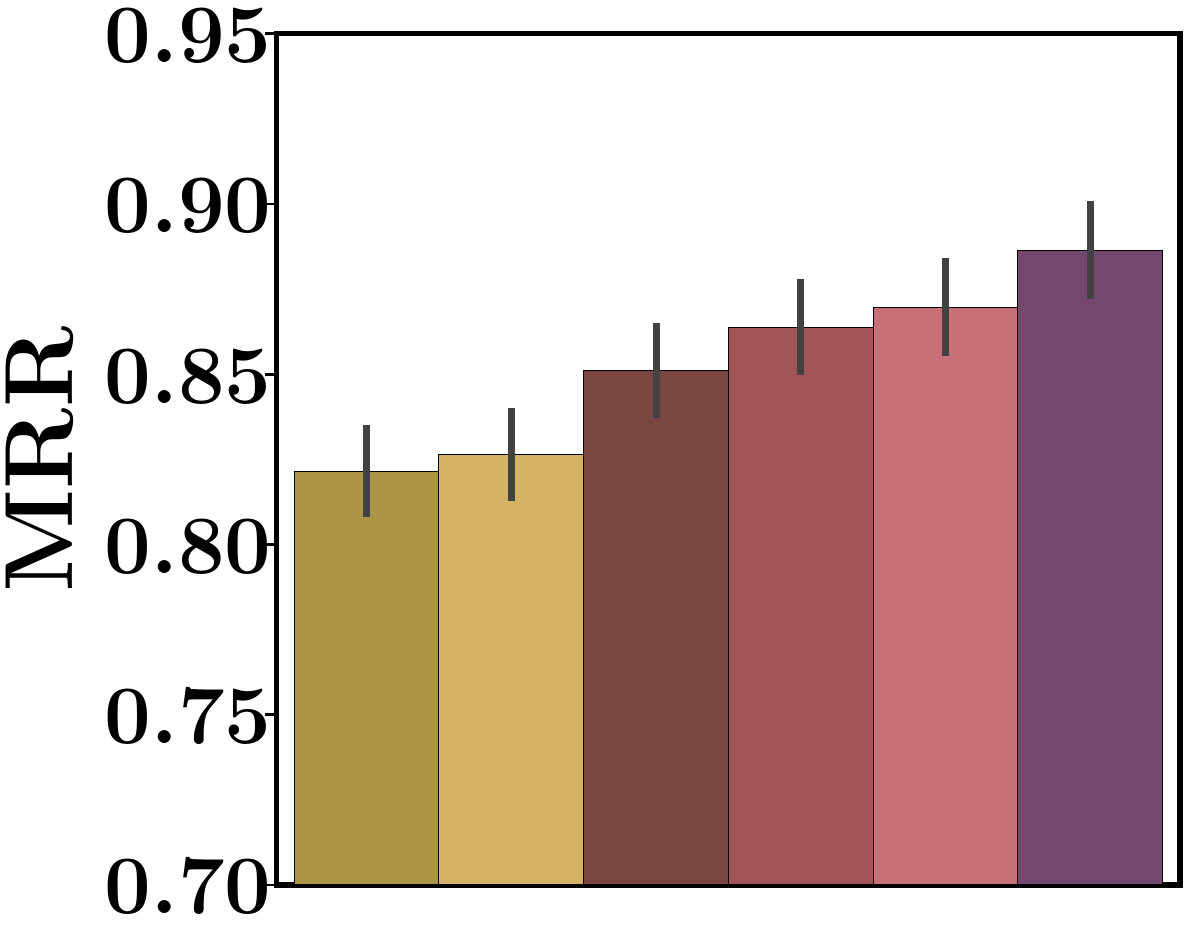}}

{\includegraphics[width=0.8\linewidth]{Figures/abs_legend.pdf}}
\vspace{-2mm}
\caption{Recommendation performance of \ourm and its variants in terms of mean reciprocal rank (MRR).}
\vspace{-2mm}
\label{fig:mrr}
\end{figure}

In addition, we report the results in terms of mean reciprocal rank (MRR) for \ourm and the best performing baselines, \ie, SASRec and TiSRec in Figure~\ref{fig:mrr}. The results across MRR show a similar trend with \ourm easily outperforming other approaches across both datasets. Interestingly, here we note that the performance difference between the baselines SASRec and TiSRec drops, \ie, the performance is similar without any significant differences.

\subsection{App and Location Prediction Category} \label{rq2}
\begin{figure}[t]
\centering
\subfloat[ST (App)]
{\includegraphics[height=2.8cm]{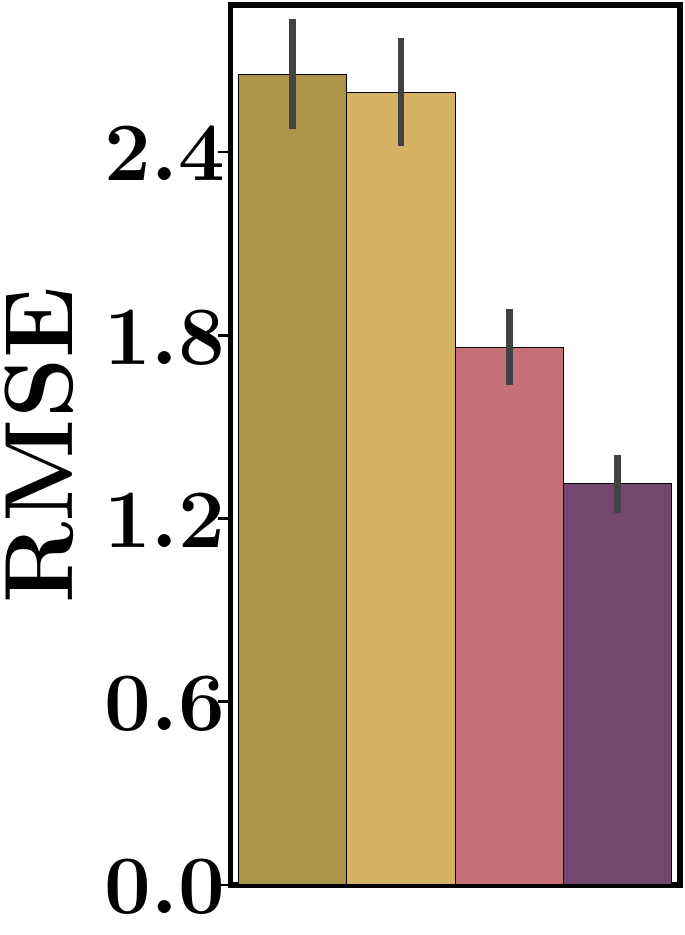}}
\hfill
\subfloat[TD (App)]
{\includegraphics[height=2.8cm]{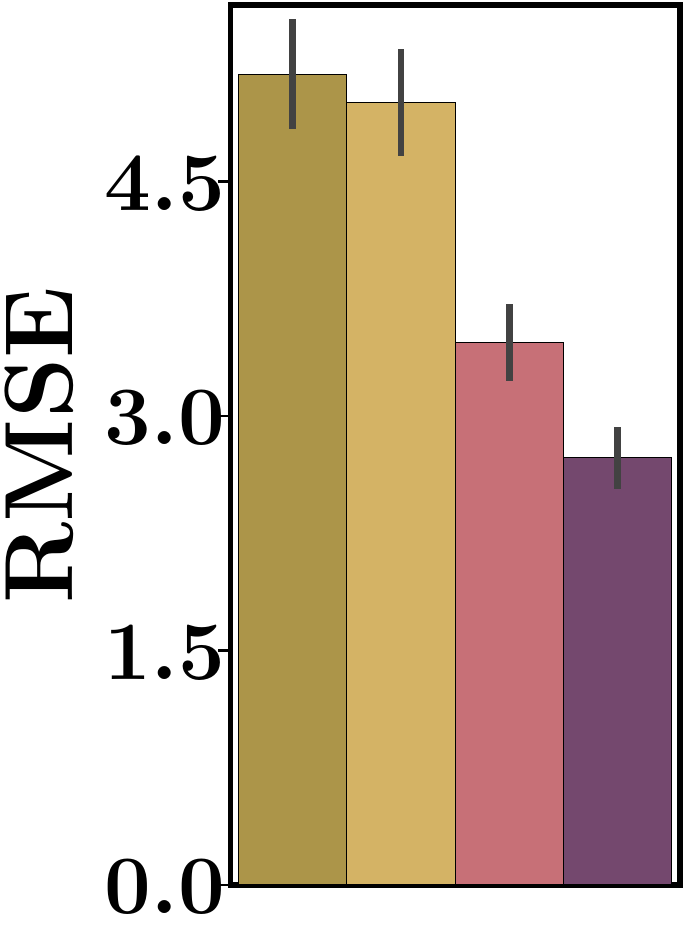}}
\hfill
\subfloat[ST (POI)]
{\includegraphics[height=2.8cm]{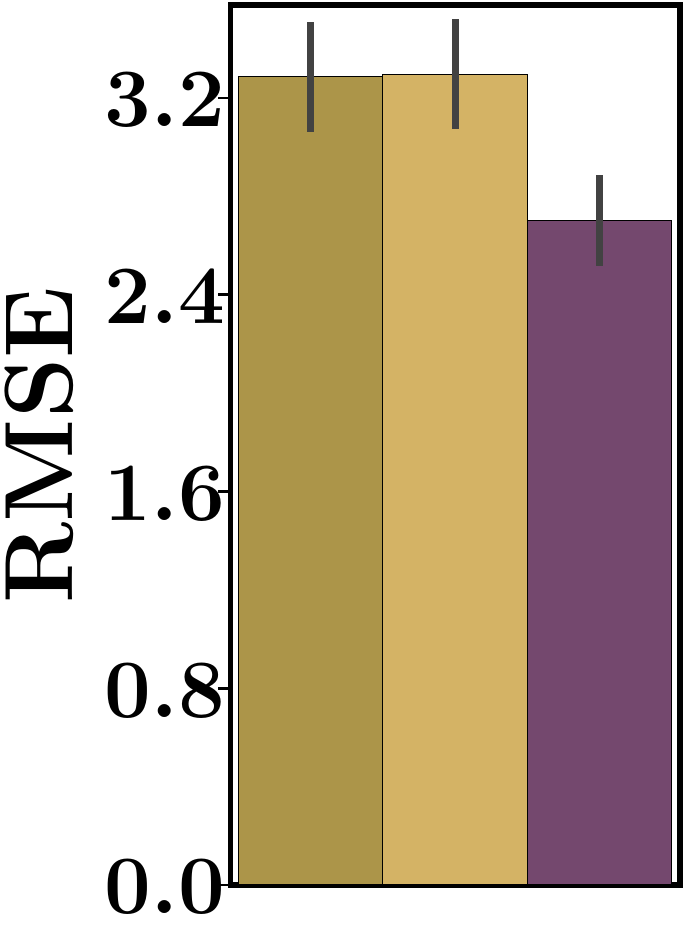}}
\hfill
\subfloat[TD (POI)]
{\includegraphics[height=2.8cm]{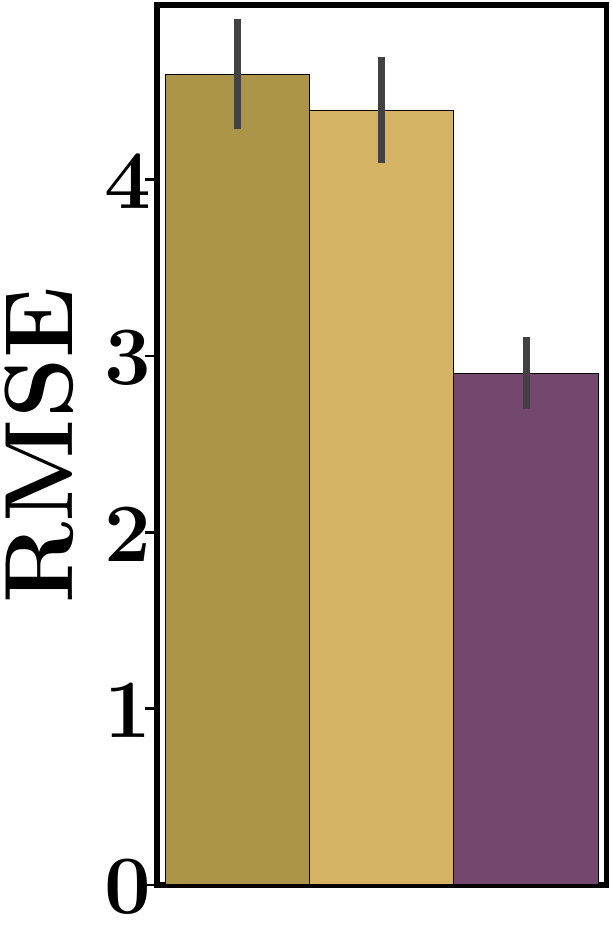}}

{\includegraphics[width=\linewidth]{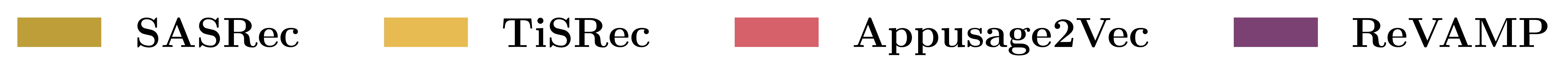}}
\vspace{-4mm}
\caption{Root-mean squared distance between the user-preference vector estimated by the model and the mean of the app- and location-category vectors of \cins.}
\vspace{-2mm}
\label{fig:category}
\end{figure}

\noindent Since our goal via \ourm is to understand the smartphone activity of a user and correlate it with her mobile trajectories. Therefore, we perform an additional experiment to evaluate how effectively is \ourm able to predict the app- and the location category for the next user \cin. We also introduce an additional state-of-the-art smartphone-activity modeling baseline, Appusage2Vec~\cite{appusage2vec} which considers the category of the app and the time spent on the app by the user to learn an app-preference embedding of a user. We also compare with the state-of-the-art transformer-based models -- SASRec~\cite{sasrec} and TiSRec~\cite{tisasrec}. For an even comparison, we rank the models using the root-mean-squared (RMS) distance between the final user preference embedding obtained after learning on $N$ consecutive events of a user and the \textit{mean} of location and category embeddings of the $N+1$ event in the sequence. Accordingly, we also modify the architectures of SASRec and TiSRec to predict user affinity across the location and app category affinities. From the results in Figure~\ref{fig:category}, we make the following observations: \begin{inparaenum}[(i)] \item  \ourm easily outperforms all other baselines for both apps and location category prediction. This illustrates the better user-preference modeling power of \ourm over other approaches, \item For app-category prediction, Appusage2Vec also outperforms both SASRec and TiSRec even with its shallow neural architecture. However \ourm easily outperforms Appusage2Vec across both datasets.\end{inparaenum}

\subsection{Scalability of \ourm (RQ3)}
\noindent To determine the scalability of \ourm with different positional encodings -- absolute and relative, we present the epoch-wise time taken for training \ourm in Table~\ref{tab:scale}. Note that these running times exclude the time for pre-processing where we calculate the inter-event app and location category-based differences. We note that the run-time of \ourm is linear with the number of users and secondly, even for a large-scale dataset, like \tdk, we can optimize all parameters within 170 minutes, and thus, in range for designing recommender systems. 

In addition, we report the training times for different subsets of data in Table~\ref{tab:subset}. Specifically, we show results where we use 40\%, 60\%, and 80\% of all users in the dataset. These users are selected randomly among all users in the complete dataset and the training-test sets are modified accordingly. All other parameters are the same as before. From the results, we make the following observations: (i) the run times increase linearly as per the subset of users; and (ii) the training times of \ourm are well within the acceptable range for practical deployment.

\begin{table}[t]
\centering
\caption{Run-time statistics of training \ourm in minutes on a 32GB Tesla V100 GPU with 256 batch-size.}
\vspace{-2mm}
\resizebox{\columnwidth}{!}{
\begin{tabular}{lrrrrrr}
\toprule
\textbf{Epochs} & \textbf{20} & \textbf{60} & \textbf{100} & \textbf{160} & \textbf{200} & \textbf{300}\\
\midrule
\tel & 2.43 & 6.31 & 10.48 & 16.72 & 21.47 & 32.01\\
\tdk & 11.39 & 33.73 & 56.12 & 89.81 & 112.35 & 168.49\\
\bottomrule
\end{tabular}
}
\label{tab:scale}
\end{table}
\begin{table}[t]
\centering
\caption{Run-time statistics of \ourm in minutes for different subsets of data -- 40\%, 60\%, and 80\%.}
\vspace{-2mm}
\begin{tabular}{l|rrr|rrr}
\toprule
 & \multicolumn{3}{c|}{\tel} & \multicolumn{3}{c}{\tdk} \\
 \hline
 & 40\% & 60\% & 80\% & 40\% & 60\% & 80\% \\
\hline
Time & 17.8 & 22.3 & 28.4 & 71.3 & 105.2 & 141.6\\
\bottomrule
\end{tabular}
\label{tab:subset}
\end{table}
\begin{figure}[t!]
\centering
\subfloat[\tel]
{\includegraphics[height=2.8cm]{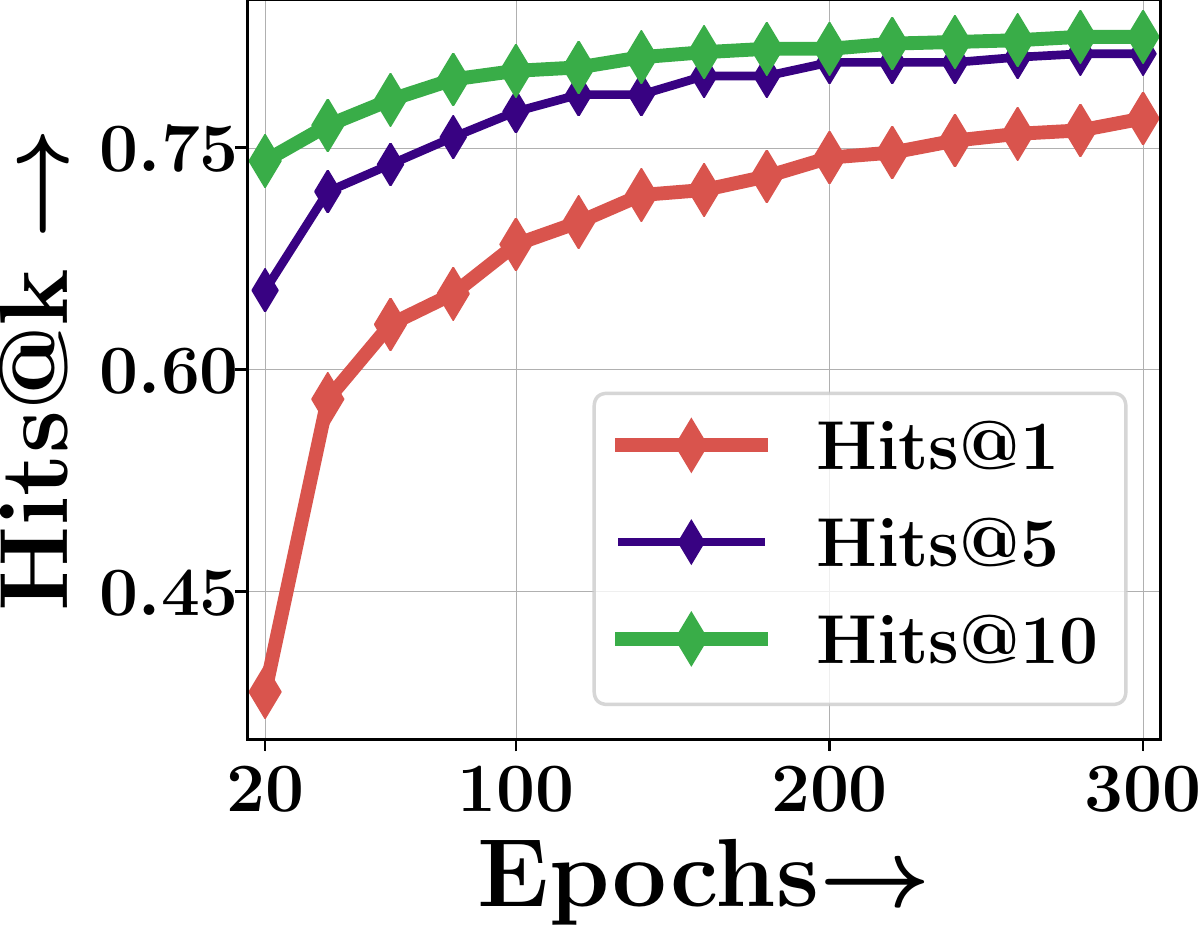}}
\hspace{0.5cm}
\subfloat[\tdk]
{\includegraphics[height=2.8cm]{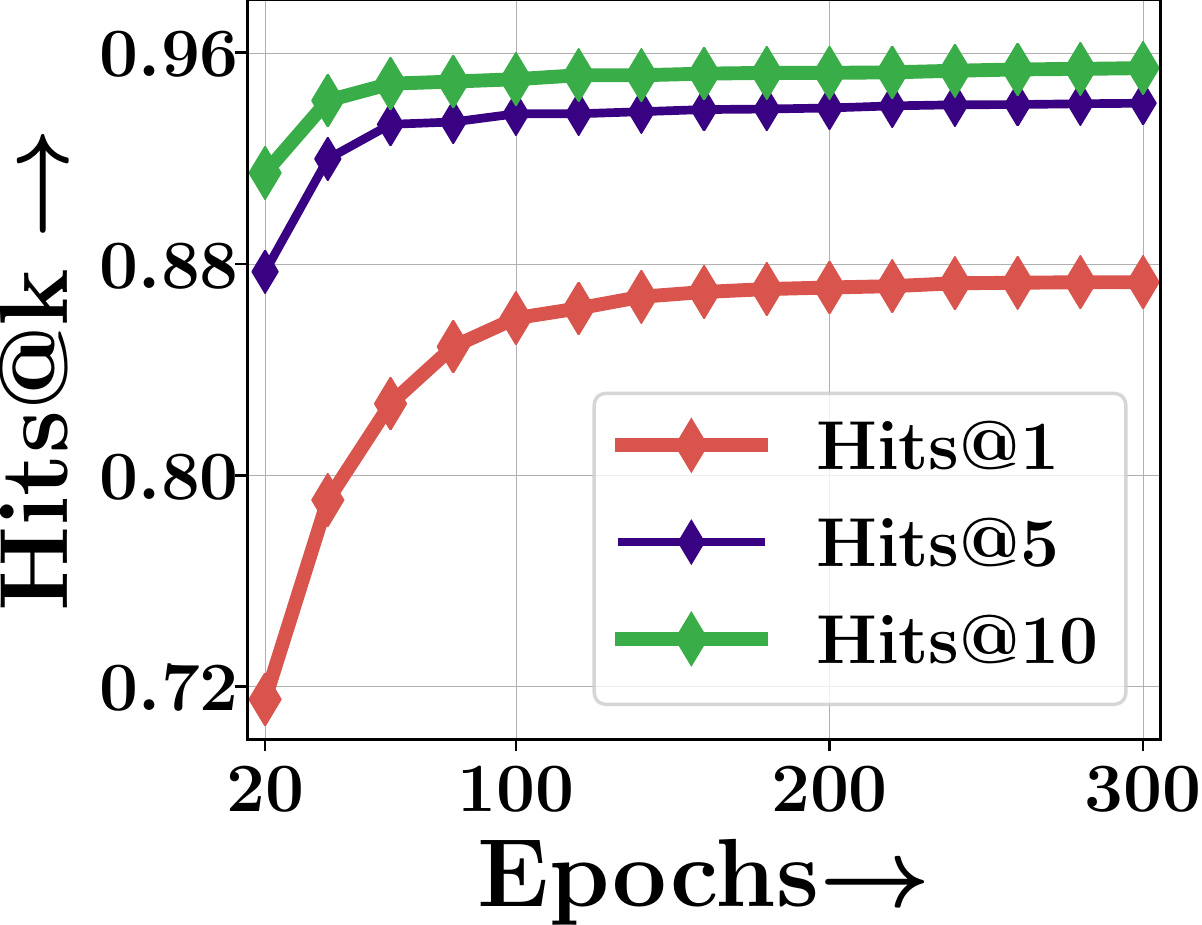}}
\vspace{-2mm}
\caption{Epoch-wise recommendation performance of \ourm for both datasets in terms of Hits@1, 5, 10.}
\vspace{-2mm}
\label{fig:runtime}
\end{figure}

\begin{figure}[t!]
\vspace{-0.2cm}
\centering
\subfloat[Across $D$]
{\includegraphics[height=2.8cm]{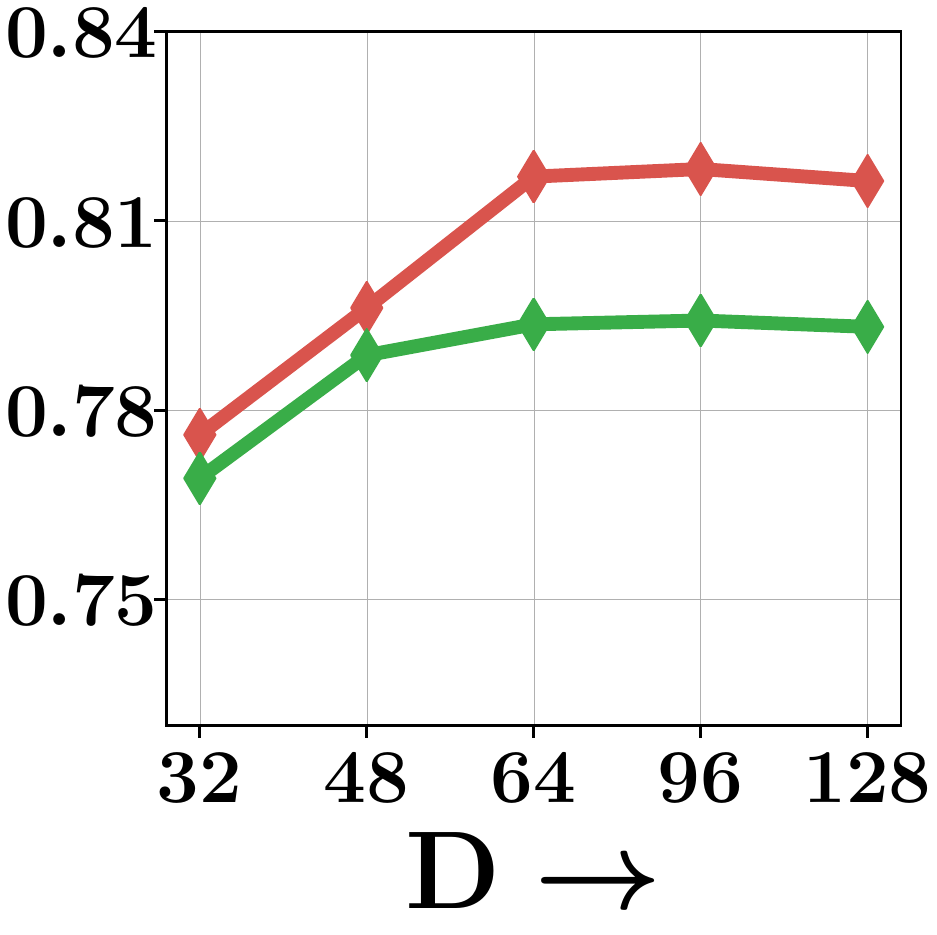}}
\hfill
\subfloat[Across $N$]
{\includegraphics[height=2.8cm]{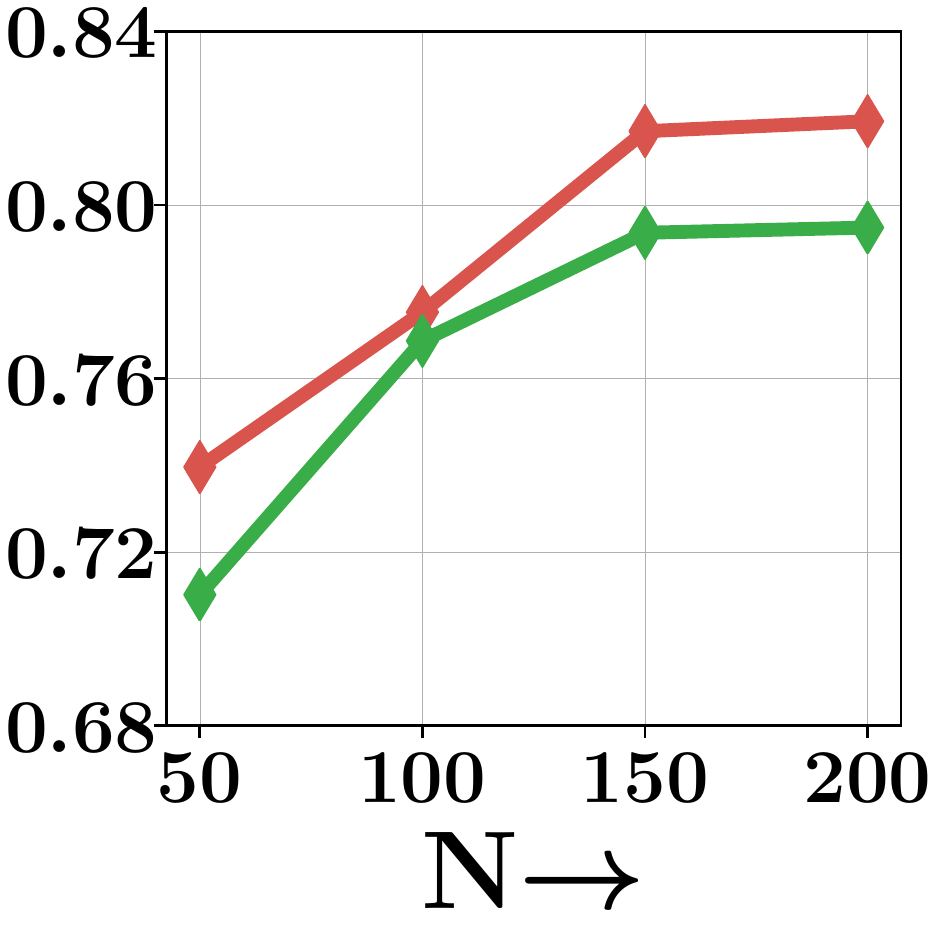}}
\hfill
\subfloat[Across $I$]
{\includegraphics[height=2.8cm]{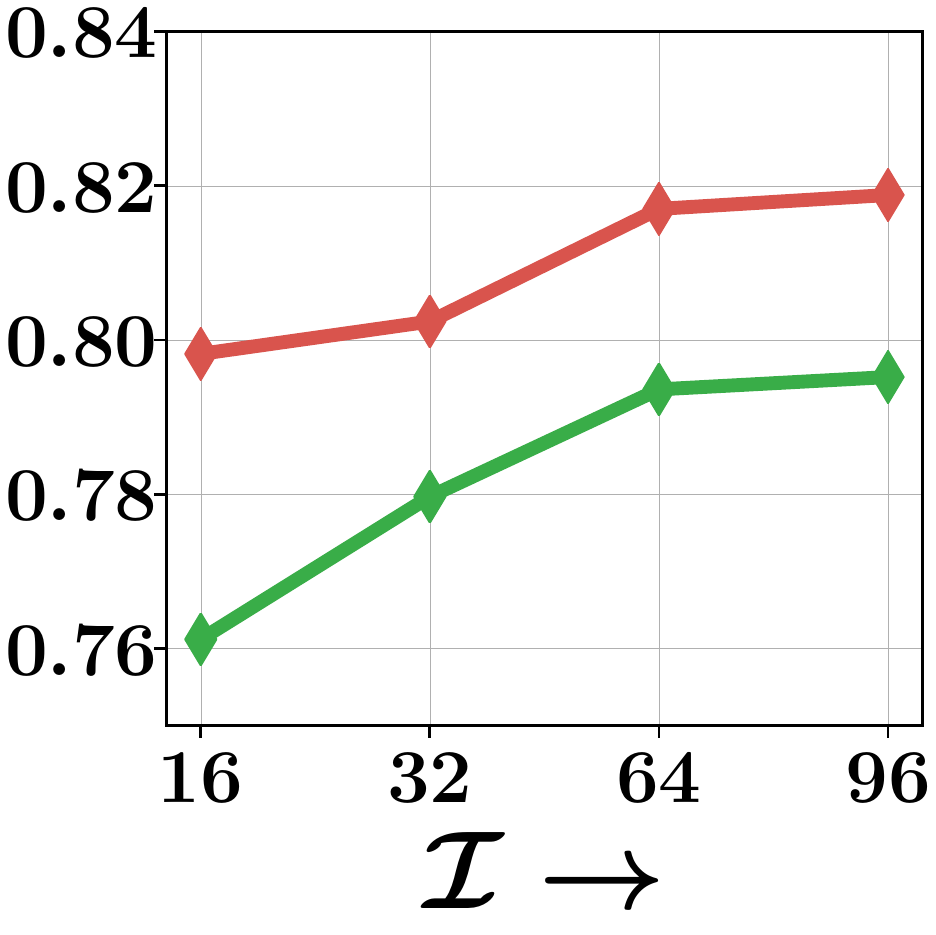}}

{\includegraphics[width=0.6\linewidth]{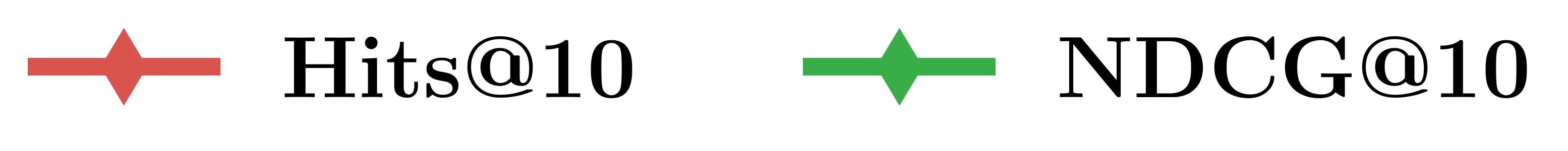}}
\vspace{-2mm}
\caption{Parameter Sensitivity for \tel dataset.}
\vspace{-2mm}
\label{fig:param_li}
\end{figure}

\xhdr{Convergence of \ourm Training}
As we propose the first-ever application of the self-attention model for smartphones and human mobility, we also perform a convergence analysis during training \ourm. To emphasize on the stability of \ourm training procedure, we plot the epoch-wise best prediction performance of \ourm across both datasets in Fig \ref{fig:runtime}. From the results, we note that despite the multi-variate nature of data and the disparate positional encodings, \ourm converges only in a few training iterations. It is also important to note that the \ourm significantly outperforms other RNN-based baselines even with limited training of 40 iterations.

\subsection{Parameter Sensitivity (RQ4)} \label{rq4}
\noindent Finally, we perform the sensitivity analysis of \ourm. The key parameters we study are (i) $D$, the dimension of embeddings; (ii) $N$, no. of latest events considered for training; and (iii) $I_{a,l,t}$, predefined normalizing constant for cosine-similarity for all relative encodings. (see Table I). In this section, we evaluate the model on NDCG@10 and Hits@10. We report the recommendation performance across different hyperparameter values for the \tel dataset and omit results for \tdk for brevity. However, we noted a similar behavior for the \tdk dataset as well. From the results in Figure~\ref{fig:param_li}, we note that as we increase the embedding dimension, $D$, the performance first increases since it leads to better modeling. However, beyond a point, the complexity of the model increases requiring more training to achieve good results, and hence we see some deterioration in performance. Next, increasing the no. of events for recommendation leads to better results before saturating at a certain point. We found $N=100$ and $N=200$ to be the optimal point across \tel and \tdk in our experiments. For normalizing constant $I$, an interesting insight is that on increasing the constant value the performance increases and later plateaus after a certain point. This could be due to saturation after a further increase in no. of distinct positional encodings.

\section{Conclusion}\label{sec:conc}
\noindent In this paper, we highlighted the drawbacks of modern POI recommender systems that ignore smartphone usage characteristics of users. Our proposed sequential POI recommendation model, called \ourm, incorporates the smartphone usage details of users while simultaneously maintaining their privacy. Inspired by the success of relative positional encodings in self-attention models, we use relative and absolute positional encodings determined by the inter-\cin variances in the smartphone app category, POI category, and time over the \cins in the sequence. Experiments over two diverse datasets show that we significantly outperform other state-of-the-art baselines. Moreover, we show the contribution of each component in the \ourm architecture and analyze the stability and the performance sensitivity across different hyperparameters. In future work, we plan to use a federated learning approach to train the model parameters with decentralized data to more effectively preserve the privacy of all users.

\section*{Acknowledgments}
\noindent This work was partially supported by a DS Chair of Artificial Intelligence fellowship to Srikanta Bedathur. 

\begin{footnotesize}
\bibliographystyle{ACM-Reference-Format}
\bibliography{refs}
\end{footnotesize}

\appendices
\vspace{-1cm}
\begin{IEEEbiography}[{\includegraphics[width=1in,height=1.25in,clip,keepaspectratio]{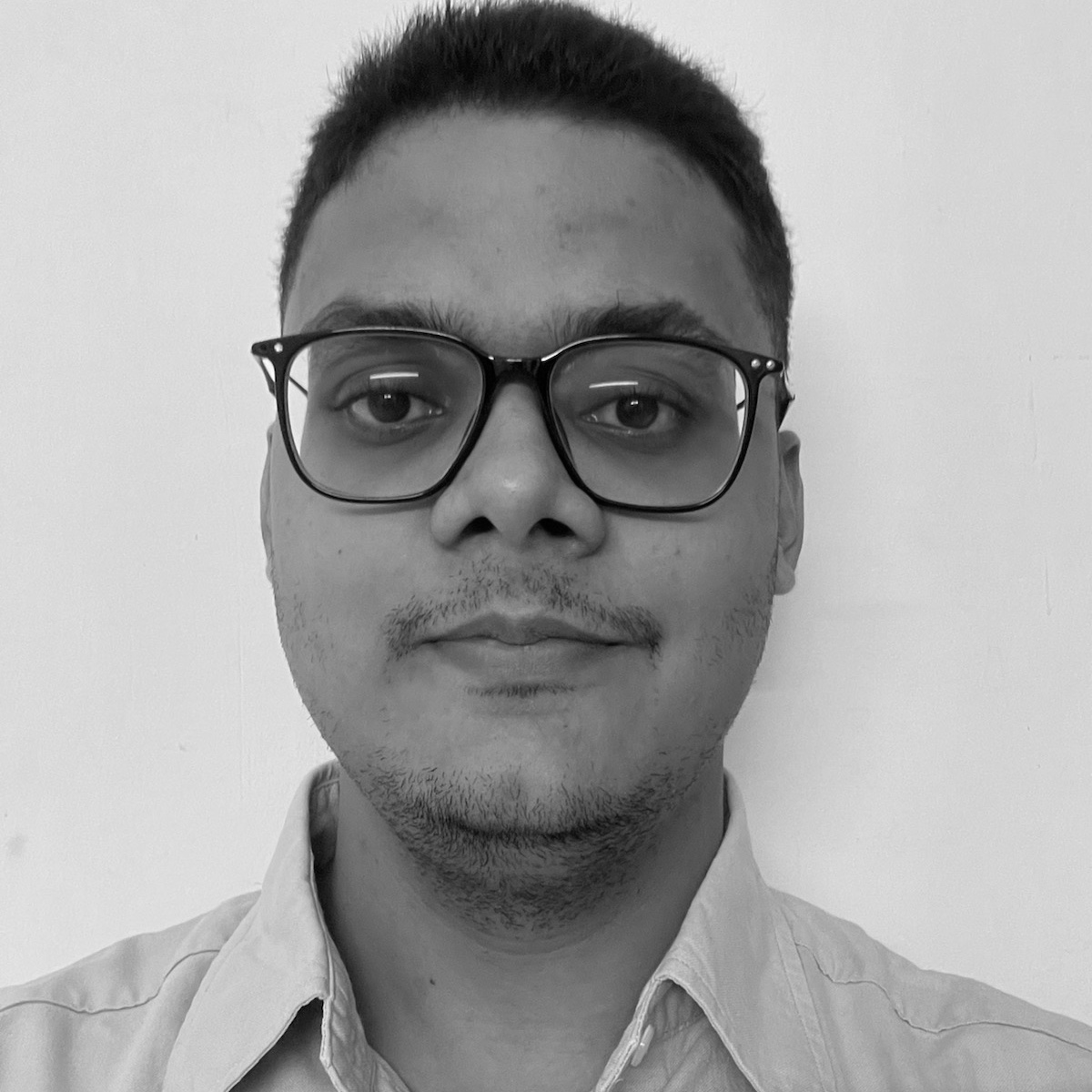}}]{Vinayak Gupta}
is a Ph.D. student under Prof. Srikanta Bedathur at the Indian Institute of Technology (IIT) Delhi. His research interests are in designing neural models for time series and graphs in the presence of missing and scarce data. He received the outstanding doctoral paper award at the conference on AI-ML Systems 2021. Previously, he completed his bachelor's in computer science from the Indian Institute of Information Technology (IIIT) Jabalpur.
\end{IEEEbiography}
\vspace{-1cm}
\begin{IEEEbiography}[{\includegraphics[width=1in,height=1.25in,clip,keepaspectratio]{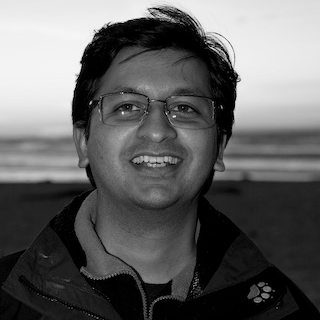}}]{Srikanta Bedathur} 
is an associate professor and DS Chair Professor of Artificial Intelligence at the Department of Computer Science and Engineering at IIT Delhi. He previously worked as a research scientist at IBM Research as part of their AI Research team, as a faculty member at IIIT Delhi, and as a senior researcher at the Max-Planck Institute for Informatics (MPII). He obtained his Ph.D. (2005) from the Indian Institute of Science, Bangalore.
\end{IEEEbiography}

\end{document}